\documentclass[runningheads]{llncs}

\usepackage{eccv}

\usepackage{eccvabbrv}

\usepackage{graphicx}
\usepackage{booktabs}
\usepackage{enumitem}
\usepackage{multirow}
\usepackage[ruled,vlined]{algorithm2e}
\usepackage[accsupp]{axessibility}  

\usepackage{hyperref}
\usepackage{orcidlink}

\usepackage[dvipsnames]{xcolor}
\usepackage{zref-base}
\usepackage{bbm}
\usepackage{amsmath}
\usepackage{amssymb}
\usepackage{booktabs}
\usepackage{esint}
\usepackage{svg}
\usepackage{zref-base}
\usepackage{atveryend}

\DeclareMathOperator*{\argmin}{arg\,min}

\makeatletter
\AfterLastShipout{%
  \if@filesw   
    \begingroup
      \zref@setcurrent{default}{\the\value{equation}}%
      \zref@wrapper@immediate{%
        \zref@labelbyprops{eq-last}{default}%
      }%
    \endgroup
  \fi
}
\makeatother

\begin{document}

\title{Towards Open-World Object-based Anomaly Detection via Self-Supervised Outlier Synthesis}
\titlerunning{Anomaly Detection via Self-Supervised Outlier Synthesis}
\author{Brian K. S. Isaac-Medina$^\star$, Yona Falinie A. Gaus$^\star$, \\ Neelanjan Bhowmik$^\star$, Toby P. Breckon$^{\star, \dagger}$}
\authorrunning{B.K.S. Isaac-Medina et al.}
\institute{Department of \{$^\star$Computer Science, $^\dagger$Engineering\}, Durham University, UK}
\maketitle
\vspace{-0.8cm}
\begin{abstract}

Object detection is a pivotal task in computer vision that has received significant attention in previous years. Nonetheless, the capability of a detector to localise objects out of the training distribution remains unexplored. Whilst recent approaches in object-level out-of-distribution (OoD) detection heavily rely on class labels, such approaches contradict truly open-world scenarios where the class distribution is often unknown. In this context, anomaly detection focuses on detecting unseen instances rather than classifying detections as OoD. This work aims to bridge this gap by leveraging an open-world object detector and an OoD detector via virtual outlier synthesis. This is achieved by using the detector backbone features to first learn object pseudo-classes via self-supervision. These pseudo-classes serve as the basis for class-conditional virtual outlier sampling of anomalous features that are classified by an OoD head. Our approach empowers our overall object detector architecture to learn anomaly-aware feature representations without relying on class labels, hence enabling truly open-world object anomaly detection. Empirical validation of our approach demonstrates its effectiveness across diverse datasets encompassing various imaging modalities (visible, infrared, and X-ray). Moreover, our method establishes state-of-the-art performance on object-level anomaly detection, achieving an average recall score improvement of over $5.4$\% for natural images and $23.5$\% for a security X-ray dataset compared to the current approaches. In addition, our method detects anomalies in datasets where current approaches fail. Code available at \url{https://github.com/KostadinovShalon/oln-ssos}.

\end{abstract}    
\section{Introduction} \label{sec:intro}

Anomaly detection plays a crucial role in identifying deviations from the norm in various applications
such as industrial inspection \cite{roth2022towards, deng2022anomaly,lee2022cfa} and video surveillance \cite{li2013anomaly, lu2013abnormal, luo2017revisit,gaus2023region}. In general, anomaly detection addresses an aspect of the open set problem in computer vision - whilst normality in terms of the appearance and behaviour of objects within the scene can be bounded, conversely, the set of possible anomalous occurrences is unbounded. Anomalous events rarely occur as compared to normal activities, which in itself results in the commonplace dataset challenge of anomaly detection - the availability of \textit{abnormal} (anomalous) samples is limited in both volume and variety. This in itself leads to a naturally imbalanced dataset distribution for any real-world anomaly detection problem. A common approach is to learn a model of the normal (non-anomalous) data distribution from the abundance of normal sample training data available and then detect anomalies as outliers in a semi-supervised manner \cite{akcay2018ganomaly, barker2021panda, roth2022towards}.
\begin{figure*}[t!]
\centering
\subfloat{\includegraphics[width=\linewidth]{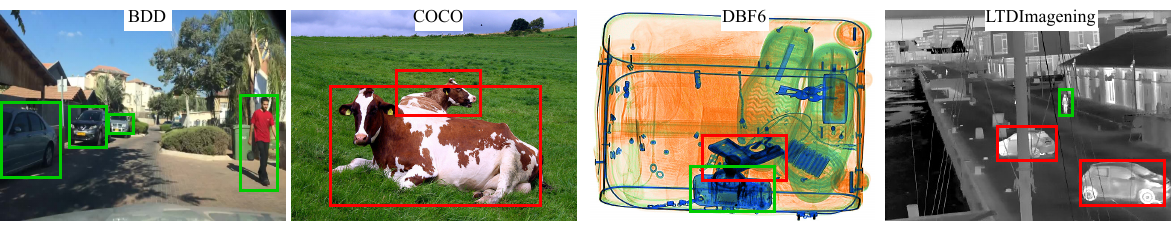}}
\caption{Exemplar of in-distribution (\textcolor{green}{green}) and out-of-distribution (\textcolor{red}{red}) objects from across four diverse benchmark datasets with various imaging modalities.}  
\label{fig:intro}
\end{figure*}

In this context, out-of-distribution (OoD) detection identifies instances from unknown (unseen) classes by training only on in-distribution data. For instance, works of \cite{liu2020energy, kumar2023normalizing} approach OoD detection by measuring the joint probability of a sample coming from one of the training classes using the free energy of the sample, subsequently identifying never-seen-before (outlier) objects. Whilst this approach has proven effective, its application within the object detection framework faces the challenge of localising unseen object categories. For example, a self-driving vehicle may encounter wild animals \cite{du2022vos}, an X-ray security screening system may identify prohibited items \cite{akcay2018ganomaly} or unknown vehicles in surveillance systems \cite{gaus2023region} (\cref{fig:intro}). These anomalous variations can vary from the visually obvious to the very subtle in the object \cite{barker2021panda} and, while current OoD approaches can classify them as anomalies, the detector must first localise them within the image. This challenge serves as motivation for this work, differentiating object-based anomaly detection from out-of-distribution detection by defining it as the joint task of localising unseen objects and identifying them as anomalous.

This work proposes an object-based anomaly detection framework that leverages an open-world object detector (OWOD), which localises unseen objects without prior class supervision. Whilst classic object detectors \cite{ren2015faster, carion2020end, wang2023yolov7} are focused on detecting objects from a set of known categories, open world object detectors \cite{kim2021oln,gupta2022ow,Zohar_2023_CVPR} naturally capture both known and unknown objects. However, the lack of classification in such OWOD subsequently disables the use of secondary class-based OoD detectors. To overcome this challenge, our method learns object pseudo-labels in a self-supervised manner consisting of alternating deep feature clustering and neural network-based prediction similar to DeepCluster \cite{caron2018deep}. We cluster the object features from the detector backbone, pooled by a RoIAlign layer \cite{he2017mask}, before each training epoch such that the resulting assignments from the clustering are used as ground truth labels to train a classifier head. Following the virtual outlier synthesis (VOS) methodology \cite{du2022vos}, class-conditional Gaussian distributions are subsequently learnt in the object feature space, which are used to sample virtual outliers from low-likelihood regions. Both in-distribution objects and virtual outliers are finally used to train a normal vs. abnormal classifier head. Since virtual outliers are sampled from the pseudo-label object distribution, we call this approach as Self-Supervised Outlier Synthesis (SSOS). 
To the best of our knowledge, we are the first to use self-supervised pseudo-classification on object instances for energy-based OoD detection, enabling object-level class-agnostic anomaly detection.
As we use the object localisation network (OLN) \cite{kim2021oln} as our OWOD, we denote our method as \textbf{OLN-SSOS}. We evaluate over a diverse set of imaging modalities and applications, 
successfully detecting anomalous instances across a wide variety of contexts. In summary, our main contributions are:
\begin{itemize}[noitemsep,topsep=0pt,leftmargin=*]
  \item \textbf{the first class-agnostic, end-to-end object-based anomaly detection architecture} that learns object pseudo-labels to fit class-conditional Gaussian distributions in the object feature space of an OWOD, thus enabling energy-based OoD detection; our approach uses self-supervised outlier synthesis (SSOS) to identify anomalies not in the training set.
  \item \textbf{state-of-the-art anomaly detection performance  across four diverse benchmark  datasets}, namely BDD100K/COCO \cite{yu2020bdd100k}, LTDImaging \cite{nikolov2021seasons}, \linebreak SIXRay10 \cite{miao2019sixray} and DBF6 \cite{akcay18architectures}, and competitive performance on the \linebreak VOC/COCO \cite{lin2014microsoft, everingham2010pascal} benchmark. Our method achieves an average recall improvement of $5.4\%$ for the BDD100k/COCO, $23.5\%$ for DBF6, and establishes the state-of-the-art for SIXRay10 and LTDImaging, where current OoD approaches fail. In addition, it achieves an average recall of $17.8\%$ (vs. $20.6\%$ using VOS \cite{du2022vos}) without class supervision.
  \item \textbf{qualitative analysis} illustrating that our architecture can jointly localise previously unseen objects within an image and classify them as anomalous, whilst other methods only identify OoD objects that are similar to the in-distribution dataset and can hence be localised by the class-based object detector (\eg, animals and vehicles are present in the training and test datasets as super-classes, with a subset of intra-class anomalous instances therein). 
 \item \textbf{supporting ablation studies} illustrating the impact of our methodological design choices, such as the number of clusters or the use of instance masks.
\end{itemize}

\section{Literature Review} \label{sec:lr}
The terminology of anomaly detection, outlier detection and out-of-distribution (OoD) detection are largely used interchangeably in the literature to describe tasks whereby the primary goal is to model the norm of a given problem domain and hence detect (outlier) deviations from that model. In general, anomaly (or outlier) detection (\cref{subsec:lr_ad}) operates under the assumption that data sample availability is highly biased towards normal classes whilst inadequate distribution coverage exists for other (abnormal) classes which may be unbounded in nature \cite{gaus2023region}. By contrast, OoD detection (\cref{subsec:lr_OoD}) leverages a closed set problem whereby in-distribution samples belong to one of a predefined set of class labels, and hence outliers are objects that do not fit into any of those category labels.

\subsection{Anomaly Detection} \label{subsec:lr_ad}


The approaches for detecting anomalies in images fall into three main categories: feature embedding, reconstruction-based, and streaming-based approaches. \linebreak Within feature embedding techniques, well-known methods include memory bank \cite{roth2022towards,cohen2020sub,defard2021padim,lee2022cfa}, knowledge distillation \cite{bergmann2020uninformed, salehi2021multiresolution,deng2022anomaly}, normalising flow networks \cite{gudovskiy2022cflow, yu2021fastflow,rudolph2021same}, and one-class classification strategies \cite{ruff2018deep, yi2020patch, zaheer2020old, sabokrou2018adversarially}. Among these, the memory bank approach, exemplified by PatchCore \cite{roth2022towards}, followed by SPADE \cite{cohen2020sub}, Padim  \cite{defard2021padim}, and CFA \cite{lee2022cfa}, stands out for their effectiveness. These methods 
extract features from all normal images and store them in a memory bank. During testing, image features are matched against the normal features stored in the memory. However, the effectiveness of these methods heavily relies on the completeness of the memory, which requires a vast collection of normal images to fully capture the normal pattern. Additionally, the size of the memory is often tied to the size of the dataset or the dimensions of the images, rendering these approaches impractical for scenarios involving large or high-resolution datasets. 

Reconstruction-based methods address anomaly detection at pixel level by typically employing autoencoders \cite{shi2021unsupervised,liu2020towards, bergmann2018improving} or generators \cite{akcay2018ganomaly,yan2021learning, zavrtanik2021draem} to encode and decode the input normal images, indirectly learning the distribution of normal images through the process of reconstruction. 
While these algorithms deliver good results, they encounter difficulties with objects of complex textures and structures. Consequently, they are prone to reconstruction errors, compromising their ability to differentiate between normal and anomalous instances.

Temporal streaming-based approaches are mostly applied in video clips \cite{ionescu2019object,liu2018future,morais2019learning,nguyen2019anomaly,park2020learning,sabokrou2018adversarially} where the task primarily involves detecting unusual events or behaviours within normal events. This is primarily achieved by analysing object trajectories \cite{li2013visual, roy2018road,roy2019adversarially} and motion characteristics  \cite{liu2018future,ravanbakhsh2017abnormal,nguyen2019anomaly,wang2021video}. For instance, Roy et al. \cite{roy2018road,roy2019adversarially} incorporate deep autoencoders to model the trajectories of normal events, subsequently identifying any abnormal trajectories as outliers. On the other hand, the work of \cite{liu2018future} introduces an optical flow loss as a motion constraint during training. In contrast, the works of \cite{ravanbakhsh2017abnormal,nguyen2019anomaly} focus on learning motion by predicting the optical flow of the current frame. Additionally, the works of \cite{wang2021video,gaus2023region} use optical flow information to guide frame prediction, where motion knowledge is used to discriminate between normal and abnormal frames.

\subsection{OoD Detection}
\label{subsec:lr_OoD}
Earlier works on out-of-distribution detection (OoD) focus on the use of generative models as a means to model the in-distribution data classes \cite{goodfellow2014generative,lee2017training,vernekar2019out,sricharan2018building}. Whilst such methods give satisfactory performance, this often drops off with increased dataset diversity and image fidelity resulting in the more recent advent of feature-based synthetic outlier generation techniques in the OoD space. 

Du et al. \cite{du2022vos} introduce virtual outlier synthesis (VOS) for OoD, employing inlier features to fit class-conditional Gaussian distributions and sampling OoD features from low likelihood regions of these distributions. Under a similar approach, \cite{du2022unknown}incorporate unknown-aware knowledge from auxiliary videos to effectively improve the performance of distinguishing OoD objects. Kumar et al. \cite{kumar2023normalizing} argue that synthesising outlier features from class-wise low-likelihood regions does not ensure that these features will not overlap other class high likelihood regions. Therefore, they use an invertible normalising flow taking all in-distribution objects into a common feature space where outliers are sampled, improving over the decision boundary between in-distribution and OoD objects. 

Whilst prior work \cite{du2022vos,du2022unknown,kumar2023normalizing} 
concentrates on feature-based OoD object detectors, \cite{wilson2023safe} leverages the backbone of an object detector network, identifying that residual convolutional layers with batch normalisation are the most effective layers for identifying OoD samples. Another notable work \cite{du2022siren} slightly deviates from this previous approach of using class-conditional Gaussian distributions \cite{du2022vos,du2022unknown,kumar2023normalizing}, by utilizing von Mises-Fisher (vMF) distributions to shape the learned representation for detecting OoD objects.

Similar to our approach, the works of \cite{joseph2021towards,gupta2022ow,zhao2023revisiting,kim2021oln} use OWOD to identify both known and unknown classes by training on pseudo-labelled unknown objects while continuously acquiring updated annotations for new unseen classes. For instance, Gupta et al. \cite{gupta2022ow} introduce a Transformer-based framework with multi-scale self-attention to discriminate between (open-set) objects and background. Wu et al. \cite{wu2022uc} incorporates a two-stage object detector to classify objects into different unknown classes, while Zhao et al. \cite{zhao2023revisiting} use a more traditional approach (selective search) to correct the auto-labelled first-stage region-proposals and subsequently classify unknown instances into new classes.


Whilst these approaches exhibit good unknown object detection performance to subsequently identify anomaly/OoD occurrences, most of the aforementioned methods present several notable challenges. First, detecting abnormalities rely heavily on suitable access to real outlier data samples \cite{li2013anomaly, lu2013abnormal, sultani2018real, luo2017revisit}, or a complex generative process to synthesise such samples\cite{ionescu2019object,liu2018future,morais2019learning,nguyen2019anomaly,park2020learning,sabokrou2018adversarially}. In real-world scenarios, anomalies can vary from the visually obvious (e.g. person dressed as a clown) \cite{chan2021segmentmeifyoucan} to the very subtle (e.g. descending fog or mist due to adverse weather condition) \cite{nikolov2021seasons}. Second, all the aforementioned work explicitly relies upon existing object-wise class labels in order to detect out-of-distribution occurrences  \cite{du2022vos,du2022unknown,kumar2023normalizing,du2022siren}. This contrasts sharply with reality, where unknown (unlabelled) object classes will naturally occur and anomaly occurrences will be a rarity. As a result, incorporating a class-agnostic OWOD is a crucial step towards building reliable object-wise anomaly detection for real-world scenarios.

\section{Methodology}
\begin{figure*}[t!]
\centering
\subfloat{\includegraphics[width=0.93\linewidth]{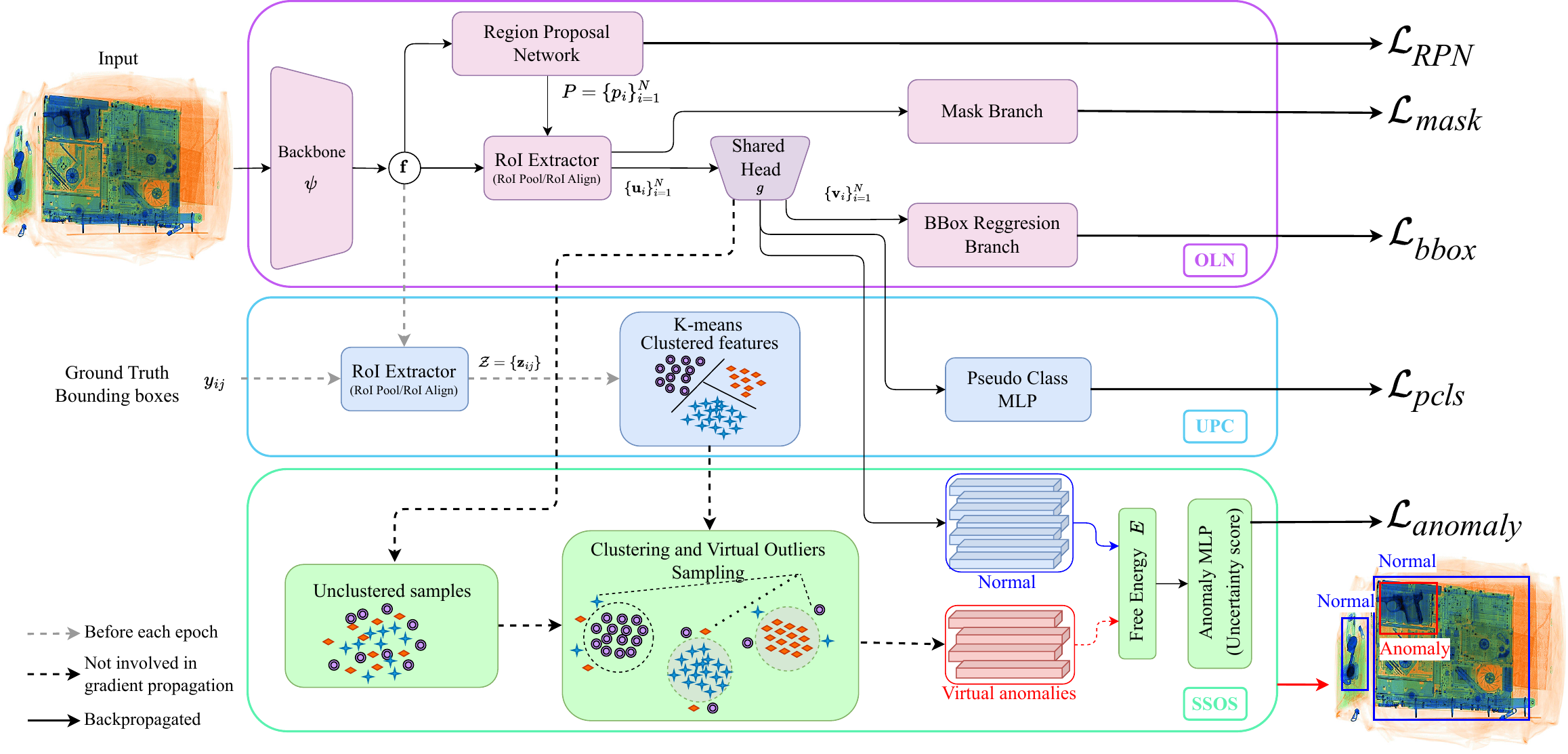}}
\caption{Our proposed architecture for open-world anomaly detection combines the object localisation network (OLN), unsupervised pseudo-classification (UPC) and anomaly detection via self-supervised outlier synthesis (SSOS).}
\label{fig:full_arch_}
\end{figure*}
\noindent
Our method consists of the combination of three architectures for different tasks that jointly enable class-agnostic object-based anomaly detection. First, an OWOD is used to detect all possible objects within the scene (\cref{sec:open-world-detector}); subsequently, an unsupervised classifier head learns to cluster object categories by pooling features from the backbone of the OWOD (\cref{sec:pseudo-classification}); and finally, a self-supervised outlier synthesis module produces a set of virtual outliers by sampling low-likelihood regions from the feature space using the learned pseudo-labels (\cref{sec:vos}). The overall architecture is illustrated in \cref{fig:full_arch_}.

\subsection{Class-agnostic Open-world Object Detection} \label{sec:open-world-detector}

In order to detect anomalies, a detector capable of localising objects belonging to unknown classes not available during training is needed. To this end, we adopt the object localisation network (OLN) architecture \cite{kim2021oln} to predict the maximal number of objects within an image. Different from standard detection, the OWOD localise \textit{all} possible objects in an image in a class-agnostic manner.

As depicted in \cref{fig:full_arch_}, the OLN consists of a region proposal network (RPN) \cite{ren2015faster}, a bounding box regression head and an optional mask branch. In this context, given an input image $\mathbf{x}$, a 2D feature map $\mathbf{f} = \psi(\mathbf{x})$ is extracted by a backbone network $\psi$. 
Subsequently, the RPN predicts a set of $N$ proposal bounding boxes $P = \left\{p_i\right\}_{i=1}^{N}, p_i \in \mathbb{R}^4$. Each object candidate $p_i$ is parameterised by a point, usually the top-left corner, and the bounding box width and height. In a departure from the classical RPN design introduced in Faster R-CNN \cite{ren2015faster} that classifies region proposals as foreground or background, the RPN in OLN regresses the centreness \cite{tian2019fcos} $c_i$ of the bounding box with the maximal overlapping ground truth bounding box as a measurement of proposal quality. Additionally, the parameters of valid proposals, \ie, those with an intersection over union (IoU) greater than a threshold, are also regressed. L1 Losses are used for the centreness and the box parameters. The RPN loss is:
\begin{equation}
    \label{eq:rpn-loss}
    \mathcal{L}_\mathit{RPN} = \frac{1}{N} \sum_{i}^N \mathrm{L1Loss}(c_i, \hat{c}_i) + \mathbbm{1}_\textit{obj}\mathrm{L1Loss}(p_i, \hat{p}_i)\,,
\end{equation}
where $\mathbbm{1}_\textit{obj}$ is $1$ if the proposal is matched against a ground truth ($0$ otherwise). Next, the proposal features $\mathbf{u}_i$ are extracted from $\mathbf{f}$ using RoIAlign \cite{he2017mask}. These features are subsequently fed into a shared network $g$ (later used at different stages) producing an object representation $\mathbf{v}_i = g(\mathbf{u}_i) \in \mathbb{R}^d$. Similarly to the RPN, a bounding box branch regresses the box quality $b_i$ and the bounding box parameters $p$ using L1 losses. The bounding box loss is:
\begin{equation} \label{eq:bbox-loss}
    \mathcal{L}_\mathit{bbox} = \frac{1}{N_v} \sum_{i}^{N_v} \mathrm{L1Loss}(b_i, \hat{b}_i) + \mathrm{L1Loss}(p_i, \hat{p}_i)\,,
\end{equation}
where $N_v$ is the number of valid proposals and the box quality $b_i$ is given by the IoU with the ground truth. An optional mask branch is added, consisting of the same mask head as in Mask R-CNN \cite{he2017mask} and a mask scoring network \cite{huang2019mask} that predicts a mask quality score $m_i$. The mask loss $\mathcal{L}_\mathit{mask}$ is hence the sum of the Mask R-CNN and Mask Scoring mask losses. During inference, the prediction score is calculated as $s_i = \sqrt{c_i b_i}$, or $s_i = \sqrt[3]{c_i b_i m_i}$ if the mask branch is used.


\subsection{Unsupervised Pseudo Classification } \label{sec:pseudo-classification}
\noindent
As it will be discussed in \cref{sec:vos}, OoD driven object-wise anomaly detection relies on the categorical distribution of the ground truth object classes at training time. However, given that our open-world object detector is class-agnostic, ground truth class labels are not available. For this reason, we perform Unsupervised Pseudo Classification (UPC) following the DeepCluster \cite{caron2018deep} methodology of learning pseudo-classes from underlying deep feature representations. In this regard, the UPC strategy consists of clustering a set of $M$ feature vectors $f_i \in \mathbb{R}^d$ into $k$ classes, using the resulting pseudo-labels to train a classifier network. Clustering is performed several times during the training process such that the clusters are re-assigned to accommodate more recently learned representations.   


While DeepCluster is implemented for image classification networks, we implement UPC by clustering object features from the backbone using a RoIAlign layer. Feature clustering is performed before each epoch via K-means trained on ground truth objects. Let $\mathcal{Z} = \left\{ \mathbf{z}_{ij}\right\}, \mathbf{z}_{ij} = \mathrm{RoIAlign}(\psi(\mathbf{x}_i), y_{ij})$ be the set of the feature representations of all ground truth bounding boxes $y_{ij}$ from all training images $\mathbf{x}_i$. K-means clustering is performed on $\mathcal{Z}$ into $K$ clusters, producing a set of cluster centres $\mathbf{w}_1, \ldots, \mathbf{w}_K$. Subsequently, each ground truth bounding box is assigned a \textit{pseudo-label} after each epoch $t$ such that:
\begin{equation}
    l^{(t)}_{ij} = \argmin_k {d(\mathbf{z}_{ij}, \mathbf{w}_k)}\,,
\end{equation}
where $l^{(t)}_{ij}$ is the bounding box $b_{ij}$ label after $t$ epochs and $d$ is the L2 distance. Finally, a multi-layer perception (MLP) predicts the pseudo-class logits $f_k$ for each object and Cross Entropy is used as the pseudo-classification loss $\mathcal{L}_\mathit{pcls}$.


\subsection{Self-Supervised Outlier Synthesis } \label{sec:vos} 

Ground truth object features and their corresponding pseudo-labels (\cref{sec:pseudo-classification}) are used to obtain class-conditional Gaussian distributions that can be used for self-supervised outlier synthesis (SSOS), thus enabling decision boundaries between inliers and outliers. We implement the virtual outlier synthesis (VOS) technique by Du \etal \cite{du2022vos}, where the class-conditional Gaussians are formed from the penultimate layer features (after the shared head, \cref{fig:full_arch_}).


A normal distribution for each object pseudo-class is constructed by having a mean $\boldsymbol{\mu}_k$ and a tied covariance $\boldsymbol{\Sigma}$ given by:
\begin{align}
    \label{eq:mean}
    \boldsymbol{\mu}_k &= \frac{1}{N_k} \sum_{i:\lambda_i = k}^{N_k} \mathbf{v}_{ij} \\
    \label{eq:std}
    \boldsymbol{\Sigma} &= \frac{1}{N} \sum_k \sum_{i:\lambda_i = k}^{N_k} (\mathbf{v}_{ij} - \boldsymbol{\mu}_k)(\mathbf{v}_{ij} - \boldsymbol{\mu}_k)^\top \,.
\end{align}
Subsequently, virtual outliers $\tilde{\mathbf{v}}$ are sampled from the normal distributions such that their probabilities are less than a value $\epsilon$. Since $\epsilon$ is unknown, an approximation is carried out by sampling several features from $\mathcal{N}(\boldsymbol{\mu}_k, \boldsymbol{\Sigma})$ and taking the less likely sample as an outlier. To differentiate between normal and anomalous objects, the free energy is used as a confidence measurement, which is given by:
\begin{equation} \label{eq:energy}
E(b_{ij}) = - \log \sum_{k=1}^{K} \exp(f_k w_k)\,,
\end{equation}
where $f_k$ are the pseudo-class logits of an object $b_{ij}$ and $w_k$ are learned weights assigning greater importance to some classes, following Du et al. \cite{du2022vos}. A greater energy indicates a more anomalous object, whereas a lower energy score suggests an object conforming to the norm. From this energy, an MLP $\phi$ is used to predict an uncertainty score $\lambda_{ij} = \phi(E(b_{ij}))$, such that normal data has greater $\lambda_{ij}$ values than outliers. During inference time, anomalies are detected by the predicted energy of objects detected by the OLN. 
With this strategy, SSOS regularises the feature representations of in-distribution objects to be compact, identifying anomalous objects by being far from all category clusters. Binary Cross Entropy is used for the classification of normal vs. anomaly, such that:
\begin{equation} \label{eq:anomaly-loss}
        \mathcal{L}_\mathit{anomaly} = \frac{1}{N_n + N_o} \left(\sum_i^{N_n} \log \left( \phi(E(b_{ij})) \right) + \sum_i^{N_o} \log \left( 1 - \phi(E(\tilde{\mathbf{v}}_i)) \right) \right) \,,
\end{equation}
where $N_n$ is the number of normal data feature vectors and $N_o$ is the number of outliers. The final loss function of our approach is thus given by:
\begin{equation}
    \label{eq:oln-vos-loss}
    \mathcal{L}_\textit{OLN-SSOS} = \mathcal{L}_\mathit{RPN} + \mathcal{L}_\mathit{bbox} + \mathcal{L}_\mathit{mask} + \alpha \mathcal{L}_\mathit{pcls} + \beta \mathcal{L}_\mathit{anomaly}\,.
\end{equation}

Feature Flow Synthesis (FFS) \cite{kumar2023normalizing}, a recent approach for outlier synthesis, uses a normalising flow function $f$ that maps the complex in-distribution features into a simpler space for feature synthesis. We also explore using this technique in our method, and call this variant OLN-FFS. Specifically, $f: \mathbb{R}^d \rightarrow \mathbb{R}^d$ is a sequence of invertible bijective functions with parameters $\theta$ that transforms the object features $\mathbf{v}_i$ into a feature space $\boldsymbol{\xi}_i = f(\mathbf{v}_i)$ such that $p(\boldsymbol{\xi}_i) \sim \mathcal{N}(\boldsymbol{0}, I)$. Virtual outliers $\tilde{\boldsymbol{\xi}}$ are sampled from this space and projected back to the object feature space via $\tilde{\mathbf{v}} = f^{-1}(\tilde{\boldsymbol{\xi}})$. During training, the log-likelihood of recovering $\mathbf{v}_i$ from $f$ is maximised by adding the negative log-likelihood loss:
\begin{equation} \label{eq:nll-loss}
    \mathcal{L}_\textit{nll} = \frac{1}{N} \sum^N_1 -\log(p_\theta (\mathbf{v}_i))\,,
\end{equation}
where $p_\theta$ is the posterior likelihood and is given by $p_\theta(\mathbf{v}_i) = p(f(\mathbf{v}_i)) \lvert \det \mathrm{J}^{f, \mathbf{v}} \rvert$, such that $\mathrm{J}^{f, \mathbf{v}_i}$ is the Jacobian matrix of $f$ with respect to $\mathbf{v}_i$. This loss is added to \cref{eq:oln-vos-loss} to form the OLN-FFS loss:
\begin{equation}
\label{eq:oln-ffs-loss}
    \mathcal{L}_\textit{OLN-FFS} = \mathcal{L}_\textit{OLN-SSOS} + \gamma \mathcal{L}_\textit{nll}\,.
\end{equation}

\section{Experimental Setup}
We evaluate OLN-SSOS and OLN-FFS on diverse datasets (\cref{subsec:datasets}) to show the effectiveness in detecting unseen anomaly objects. \cref{subsec:metric} reviews the performance metrics to evaluate our approach against the baseline methods and finally \cref{subsec:exp_setup} gives an overview of our implementation details for reproducibility.

\subsection{Datasets} \label{subsec:datasets}

In order to perform anomaly detection, datasets without object anomalies must be used for training. We evaluate diverse datasets from various image modalities (visible, infrared and X-ray). Following OoD works \cite{du2022siren, du2022vos, kumar2023normalizing}, we use the\textbf{ PASCAL-VOC 2007 and 2012} \cite{everingham2010pascal} datasets with $20$ object categories and the \textbf{Berkeley DeepDrive (BDD100K)} \cite{yu2020bdd100k} dataset with $10$ categories as ID while the OoD test sets consist on subsets of the MS-COCO \cite{lin2014microsoft} validation partition removing images containing in-distribution instances. To demonstrate the efficacy of our method in actual application scenarios, we also train our model on two X-ray security imagery datasets, \textbf{SIXray10} \cite{miao2019sixray}, a publicly available security inspection X-ray image with $5$ object class labels and \textbf{ Durham Baggage Full Image (DBF6)} \cite{akcay2018ganomaly} datasets containing $6$ object class labels, and apply a leave-one-out contraband item anomaly detection formulation (`firearms' in SIXRay10, `firearms' + `firearm parts' in DBF6) to construct the in-distribution/OoD training and testing data partitions. Finally, we also use the publicly available \textbf{Long-term Thermal Drift (LTD)} \cite{nikolov2021seasons} dataset, similarly applying a leave-one-out strategy (vehicle). A summary of the composition of these dataset formulations is presented in Supplementary Material.


\subsection{Evaluation Metrics}\label{subsec:metric}
OoD detectors focus on evaluating how accurately the predicted detections in the OoD dataset are flagged as outliers while keeping the in-distribution dataset detections with low false positive anomalies. However, this approach does not account for the anomalies recall. Motivated by this, we report class-agnostic MS-COCO \cite{lin2014microsoft} detection metrics to investigate the localisation performance. Since our method leverages an open-world detector, only the single-class average recall (AR) metrics are reported, specifically AR@10 ($10$ detections), AR@100 ($100$ detections), AR@S (small objects), AR@M (medium objects) and AR@L (large objects). Following convention, OoD detection performance is reported considering detections with an uncertainty score below a threshold such that $95\%$ of in-distribution detected objects are above it (\ie, they are flagged as normal). Following Du et al. \cite{du2022vos}, only in-distribution detections with a confidence score greater than an optimum threshold (that maximises the F1 score) are considered.


\subsection{Implementation details}\label{subsec:exp_setup}

The OWOD sub-network of OLN-SSOS and OLN-FFS is implemented following the original OLN architecture \cite{kim2021oln}, \ie, a Faster-RCNN \cite{ren2015faster} (or Mask RCNN \cite{he2017mask} for instance segmentation) detector with a ResNet-50 \cite{he2016deep} backbone pre-trained on the ImageNet \cite{krizhevsky2017imagenet} and with no classification heads. In this sense, all ground truth class labels are ignored in order to account for learned pseudo-classes. UPC is carried out before each epoch using the ground truth object features extracted from the backbone (\cref{sec:pseudo-classification})  using a RoIAlign layer with a $3\times3$ output size and $256$ channels. These features are flattened, creating a $2,304$ vector representation of each bounding box. Pseudo-labels are obtained using the mini-batch $k$-means implementation of Sculley \cite{minibatchkmeans}, using the resulting cluster centres as initialisation for the next epoch re-clustering. These pseudo-labels are used as ground truth classes to train the pseudo-label classifier (no background class is added), with a loss weight of $1$ ($\alpha$ in \cref{eq:oln-vos-loss}). The corresponding SSOS and FFS implementations follow the original settings, such that the anomaly classification module in \cref{fig:full_arch_} consist of a two-layer MLP with a ReLU activation and $512$ hidden dimensions. We use a loss weight of $0.1$ ($\beta$ in \cref{eq:oln-vos-loss}). OLN-FFS models use an \textit{nll} loss weight of $1\times10^{-4}$ ($\gamma$ in \cref{eq:oln-ffs-loss}). For OLN-SSOS, outliers are chosen as the least confident out of $10,000$ class-conditional samples, while for OLN-FFS the samples are reduced to 300.  We investigate the impact of the number of pseudo-labels, as well as the number of outlier samples in the ablation studies. Since DBF6 is the only dataset with available instance masks, we include variants using OLN-Mask \cite{kim2021oln}. We homogenise all implementations under the MMDetection \cite{mmdetection} framework (VOS and FFS are implemented using Detectron2 \cite{wu2019detectron2}). The training regime is detailed in Supplementary Material. 

Our proposed methods are compared against object-based OoD detectors SIREN \cite{du2022siren}, VOS \cite{du2022vos} and FFS \cite{kumar2023normalizing}. SIREN and VOS use a ResNet-50 backbone while FFS is trained with a RegNetX-50\cite{Radosavovic_2020_CVPR}. For VOC and BDD in-distribution datasets, the original settings are used for the baselines. For DBF6, SIXRay10 and LTDImaging, all baseline methods are trained using similar configurations as in VOS, \ie, minimum image size of $800$ (except for LTDImaging, which is trained using an image size of 384), $10,000$ samples for virtual outlier synthesis and similar training recipe consisting of $18$ epochs with a learning rate of $0.02$ decaying by a factor of $10$ in epochs $12$ and $16$. Following the original works, outlier synthesis starts at epoch $12$ for VOS and FFS. All experiments (our approach and the baselines) are trained using a single NVIDIA 2080Ti GPU.

\section{Results} \label{s:result}
\noindent
\cref{tab:voc_bdd_results,tab:dbf6_results,tab:sixray10_ltd_results} compare our method with state-of-the art OoD object detectors SIREN \cite{du2022siren}, VOS \cite{du2022vos} and FFS \cite{kumar2023normalizing}. While these methods are evaluated using the area under the receiver operating characteristic (AUROC) and the false positive rate at $95\%$ in-distribution recall (FPR95), we instead report AR to assess detection performance (\cref{subsec:metric}). 
Only in-distribution data is used during all training.

\begin{figure*}[t!]
\centering
\subfloat{\includegraphics[width=\linewidth]{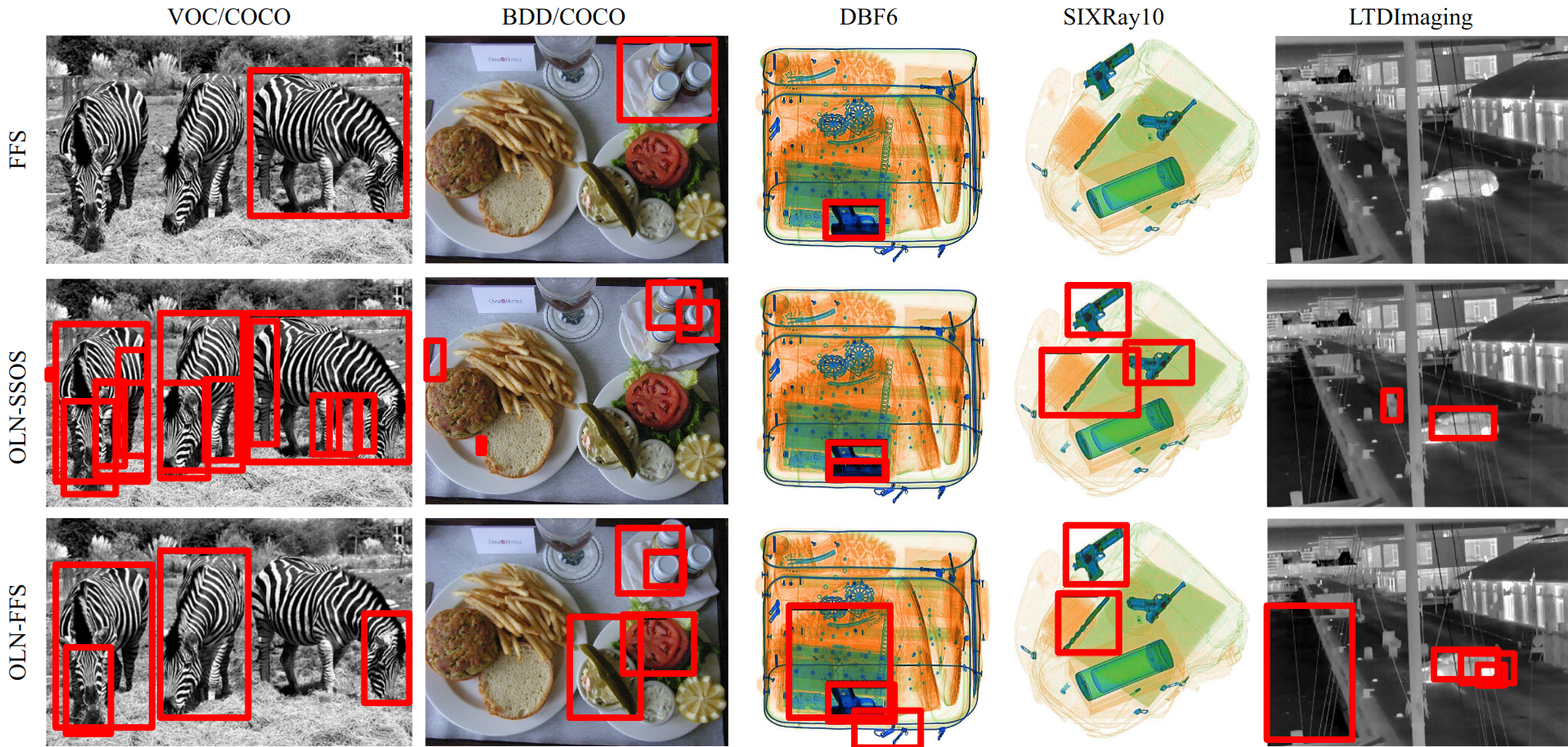}}
\caption{The qualitative results exemplify the effectiveness of our proposed approaches (OLN-SSOS/OLN-FFS) in detecting OoD/anomalous objects (in \textcolor{red}{red} bounding boxes).}  
\label{fig:qualitative_results}
\end{figure*}

\begin{table}[t]
\centering
\caption{Anomaly detection metrics for PASCAL VOC and BDD datasets.}
\label{tab:voc_bdd_results}
\resizebox{\linewidth}{!}{%
\renewcommand{\arraystretch}{1}
\begin{tabular}{|c|l|cccccc|cccccc|}
\hline
\multicolumn{2}{|c|}{} & \multicolumn{6}{|c|}{In-distribution Test Set} & \multicolumn{6}{|c|}{COCO OoD Test Set} \\ \hline
 & \textbf{Method} & \textbf{AR@1} & \textbf{AR@10} & \textbf{AR@100} & \textbf{AR@S} & \textbf{AR@M} & \textbf{AR@L} & \textbf{AR@1} & \textbf{AR@10} & \textbf{AR@100} & \textbf{AR@S} & \textbf{AR@M} & \textbf{AR@L} \\ \hline
\multirow{5}{*}{\rotatebox[origin=c]{90}{\scriptsize PASCAL-VOC}}&SIREN \cite{du2022siren}   & 23.9  & 56.5  & 59.7  & 37.3  & 52.5  & 68.6  & 9.0   & 19.2  & 19.9  & 4.5   & 8.8   & 35.2  \\
&VOS \cite{du2022vos}    & 23.8  & 56.3  & 59.5  & 37.0  & 52.2  & 68.6  & \textbf{9.8}   & \textbf{20.0}  & \textbf{20.6}  & 4.1   & 9.7   & \textbf{36.3}  \\
&FFS  \cite{kumar2023normalizing}   & \textbf{24.7}  & \textbf{58.1}  & 60.9  & 36.2  & 54.3  & \textbf{70.0}  & \textbf{9.8}   & 19.2  & 19.6  & \textbf{4.6}   & 9.9   & 33.8  \\
&OLN-SSOS (Ours) & 14.6  & 48.9  & 60.7  & \textbf{44.3}  & 57.6  & 66.1  & 4.3   & 11.1  & 14.8  & 2.5   & 12.2  & 21.8  \\
&OLN-FFS (Ours) & 14.9  & 49.6  & \textbf{61.3}  & 44.0  & \textbf{58.4}  & 66.7  & 3.2   & 11.2  & 17.8  & 3.9   & \textbf{19.2}  & 22.1  \\ \hline
\multirow{5}{*}{\rotatebox[origin=c]{90}{\scriptsize BDD}}&SIREN \cite{du2022siren}  & 4.6  & \textbf{32.3}  & \textbf{51.8}  & \textbf{37.6}  & \textbf{63.2}  & \textbf{85.6}  & \textbf{3.0}    & \textbf{9.2}  & 10.5  & 3.7   & 7.2   & 19.6  \\
&VOS  \cite{du2022vos}   & 4.6  & \textbf{32.3}  & 51.7  & 37.5  & \textbf{63.2}  & 85.4  & 2.6  & 8.6  & 9.9   & 3.4   & 7.0     & 18.4  \\
&FFS  \cite{kumar2023normalizing}   & 4.5  & 31.9  & 51.4  & \textbf{37.6}  & 62.6  & 84.4  & \textbf{3.0}    & 9.0    & 10.3  & 3.5   & 6.9   & 19.4  \\
&OLN-SSOS (Ours) & \textbf{4.7}    & 27.9     & 45.9     & 29.6     & 59.7     & 83.4     & 0.5    & 1.6    & 3.5     & 3.0     & 6.0     & 1.4    \\
&OLN-FFS (Ours) & 4.6  & 27.0    & 44.1  & 27.8  & 57.8  & 81.7  & 1.6  & 6.0    & \textbf{15.9}  & \textbf{4.4}   & \textbf{17.1}  & \textbf{24.4}  \\ \hline
\end{tabular}%
}
\renewcommand{\arraystretch}{1}
\end{table}
\cref{tab:voc_bdd_results} presents the anomaly detection performance when training on the VOC and BDD100k datasets.
Our method exhibits a great performance for the in-distribution PASCAL-VOC dataset despite being trained without class supervision 
(AR@100=$61.3$\% vs. $60.9\%$ for FFS) and a competitive OoD detection, with AR@100=$17.8$\% compared with $20.6\%$ for VOS. We highlight that our method does not use class labels, making it essentially unsupervised in this aspect, and further remark on the importance of such a class-agnostic focus for a technique where the prior class distribution is a strong prior that enables energy-based OoD detection \cite{du2022vos}. Learning class distribution without class supervision is challenging given the intra-class variability (i.e., there are several modalities of the person class). Considering that VOC dataset is highly unbalanced, with several person instances, our method achieves a competitive performance. Similar in-distribution performance is observed on the BDD100k dataset. However, our approach significantly outperforms the baseline for OoD detection, with a maximum AR@100 of $15.9\%$ (OLN-FFS) vs. $10.5\%$ for the best baseline method (SIREN). 
This indicates that our approach achieves stronger OoD detection performance while maintaining moderately high accuracy in the original in-distribution object detection task. Furthermore, while the original implementation of FFS has greater FPR95 and AUROC metrics \cite{kumar2023normalizing}, this does not translate into localising anomalies, as evidenced by its low AR, especially for the BDD100k dataset that contains less classes than VOC. Our method overcomes this issue by sampling outliers from pseudo-class features with a normalising flow function, effectively detecting and localising anomalies, thus achieving stronger OoD detection performance. Qualitative results are presented in \cref{fig:qualitative_results}, with FFS chosen as the baseline since it obtains better metrics overall. It is observed that the baseline struggles to localise objects, while our method correctly localises and detects anomalies. Further qualitative results are included in the Supplementary Material. 


\begin{figure*}[t!]
\centering
\subfloat{\includegraphics[width=\linewidth]{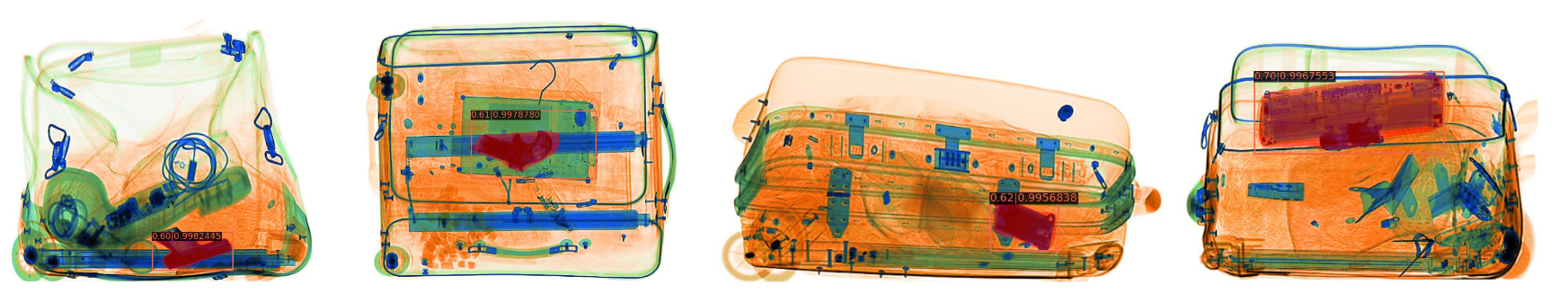}}
\caption{Qualitative results obtained from the DBF6 dataset utilising OLN-SSOS Mask.}  
\label{fig:qualitative_results_mask}
\end{figure*}

\begin{table}[t]
\centering
\caption{Anomaly detection metrics for the DBF6 dataset.}
\label{tab:dbf6_results}
\resizebox{\linewidth}{!}{%
\renewcommand{\arraystretch}{0.95}
\begin{tabular}{|l|cccccc|cccccc|}
\hline
 & \multicolumn{6}{|c|}{In-distribution Test Set} & \multicolumn{6}{|c|}{OoD Test Set} \\ \hline
\textbf{Method} & \textbf{AR@1} & \textbf{AR@10} & \textbf{AR@100} & \textbf{AR@S} & \textbf{AR@M} & \textbf{AR@L} & \textbf{AR@1} & \textbf{AR@10} & \textbf{AR@100} & \textbf{AR@S} & \textbf{AR@M} & \textbf{AR@L} \\ \hline
SIREN \cite{du2022siren}  & 47.4	&56.3&	56.4	&17.8&45.8&	86.3 &29.0&	35.1&	35.1&	0.0	&35.2&	34.0\\
VOS \cite{du2022vos}   &43.9&	54.3	&54.4&	17.6&	44.4&	82.7& 23.5	&32.8	&32.8&	0.0&	31.7&	49.1\\
FFS  \cite{kumar2023normalizing}  &\textbf{49.0}	&\textbf{56.5}	&56.5	&14.2	&45.7	&87.5 &30.0&	35.4&	35.4	&0.0	&35.1&	41.4\\ \hline
OLN-SSOS Box(Ours)  &27.7&44.2&49.1&15.2&37.0&81.9&29.4&46.1&48.8&0.0&48.3&57.4\\
OLN-FFS Box (Ours) &29.2&45.8&51.5&16.6&40.1&83.1&15.6&35.9&46.3&0.0&46.0&51.4\\ \hline

OLN-SSOS Mask (Ours) &31.3&47.0&52.0&17.0&38.9&87.3&\textbf{34.1}&\textbf{55.1}&\textbf{58.9}&0.0&\textbf{59.0}&58.6\\
OLN-FFS Mask (Ours) &32.8	&50.3&	54.8&	22.2	&42.1	&88.9& 21.7&	47.9&	57.2&	\textbf{10.0}&	57.0&	62.1\\ \hline
\end{tabular}%
}
\renewcommand{\arraystretch}{1}
\end{table}
\begin{table}[t]
\centering
\caption{Anomaly detection metrics for the SIXRay10 and LTDImaging datasets.}
\label{tab:sixray10_ltd_results}
\resizebox{\linewidth}{!}{%
\renewcommand{\arraystretch}{0.95}
\begin{tabular}{|c|l|cccccc|cccccc|}
\hline
\multicolumn{2}{|c|}{} &  \multicolumn{6}{|c|}{In-distribution Test Set} & \multicolumn{6}{|c|}{OoD Test Set} \\ \hline
& \textbf{Method} & \textbf{AR@1} & \textbf{AR@10} & \textbf{AR@100} & \textbf{AR@S} & \textbf{AR@M} & \textbf{AR@L} & \textbf{AR@1} & \textbf{AR@10} & \textbf{AR@100} & \textbf{AR@S} & \textbf{AR@M} & \textbf{AR@L} \\ \hline
\multirow{5}{*}{\rotatebox[origin=c]{90}{\scriptsize SIXRay10}}& SIREN \cite{du2022siren}  & 47.8&	63.3&	63.7&	10.0	&62.8&	64.7 & 0.8&	0.8&	0.8&	0.0	&0.0&	0.9\\
&VOS \cite{du2022vos}   &48.2	&63.6&	63.6&	0.0&	62.2&	65.0 & 0.0	&0.1&	0.1&	0.0&	0.0&	0.2\\
&FFS  \cite{kumar2023normalizing}  &\textbf{49.2}&	\textbf{65.4}&	\textbf{65.4}&	60.0&	\textbf{65.1}&	\textbf{65.8} & 0.8&	0.8&	0.8&	0.0&	0.0&	1.0\\
&OLN-SSOS (Ours) &28.0&	49.2&	55.2&	\textbf{70.0}&	56.4&	54.4& 10.7&	25.8&	35.3&	55.0&	34.2&	35.3\\
&OLN-FFS (Ours) &29.8&	50.4&	55.1&	30.0&	55.3&	55.1& \textbf{12.1}&	\textbf{27.3}&	\textbf{35.6}&	40.0&	32.0&	\textbf{36.0}\\ \hline
\multirow{5}{*}{\rotatebox[origin=c]{90}{\scriptsize LTDImaging}}&SIREN \cite{du2022siren}  &5.9&	34.2&	\textbf{52.5}&	\textbf{51.6}&	75.1&	- & 0.0&0.0&0.0&0.0&0.0&0.0\\
&VOS \cite{du2022vos}   &\textbf{6.0}	&\textbf{34.3}&	\textbf{52.5}&	\textbf{51.6}&	75.1&	-&	0.0&0.0&0.0&0.0&0.0&0.0\\
&FFS  \cite{kumar2023normalizing}  &\textbf{6.0}	&34.2&	\textbf{52.5}&	\textbf{51.6}&	\textbf{75.6}&	-&0.0&0.0&0.0&0.0&0.0&0.0\\
&OLN-SSOS (Ours) &3.6	&15.5	&17.8	&15.7	&70.9	&-& 0.0&	12.2&	\textbf{18.2}&	0.0&	\textbf{17.6}&	\textbf{62.9}\\
&OLN-FFS (Ours) &3.9	&16.8	&19.4	&17.3	&70.5	&-& 4.2&	\textbf{12.3}&	12.8&	2.6&	9.2&	50.2 \\ \hline
\end{tabular}%
}
\renewcommand{\arraystretch}{1}
\end{table}
\cref{tab:dbf6_results} presents the results on the DBF6 dataset. We report detection performance in both bounding box and mask detection, demonstrating the extension of our approach capability to instance segmentation. Our method achieves significantly superior results on the OoD test set without affecting the in-distribution performance. We highlight that incorporating mask features effectively enhances OoD detection results (AR@10=$55.1$\%, AR@100=$58.9$\%). Qualitative results in \cref{fig:qualitative_results,fig:qualitative_results_mask} show that our method detects firearm and firearm parts as anomalies. Other electronics (tablets) are also detected, which are not present in the training set, underscoring the benefits of integrating mask features in SSOS.  

\cref{tab:sixray10_ltd_results} shows the results for SIXRay10 and LTDImaging. Here it is observed that while the baseline methods can perform in-distribution detection, they catastrophically fail to perform anomaly detection, with 0\% AR. On the other hand, our approach offers in-distribution and OoD detection capabilities. For instance, OLN-SSOS achieves competitive in-distribution metrics, with AR@100=55.2\% vs. 65.4\% of FFS, while having an OoD AR@100 of 35.6\%. Similarly, while our methods have a significant drop in in-distribution performance on the LTDImaging dataset, they show great OoD detection, with an AR@100 up to 18.2\%. Ablation studies (\cref{sec:ablations}) show that the number of pseudo-labels may impact the performance, thus explaining the drop in in-distribution detection. Qualitative results in \cref{fig:qualitative_results} show that the baseline methods cannot detect the objects since they are trained to only detect in-distribution objects.

Finally, the results in \cref{tab:voc_bdd_results,tab:sixray10_ltd_results} show superior OoD performance of OLN-FFS against OLN-SSOS for the VOC/COCO, BDD/COCO and SIXRay10 datasets, results on the DBF6 and LTDImaging show a drop in performance when using FFS. In this context, the work of \cite{schirrmeister2020understanding} shows that invertible mapping, as in FFS, helps in forming high likelihood regions based on high-level object features. The OoD instances in DBF6 and LTDImaging are significantly different from the in-distribution classes such that these datasets might require stronger low-level features discrimination in order to achieve improved anomaly detection.


\begin{figure}[t]
     \centering
     \hspace{-0.34cm}
     \begin{subfigure}[b]{0.26\textwidth}
         \centering
         \includegraphics[width=\textwidth]{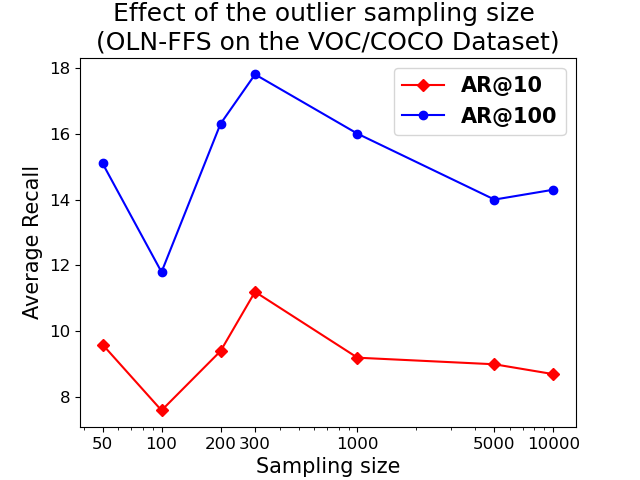}
         \caption[]{}
         \label{fig:voc_coco_ablations}
     \end{subfigure}
     \hspace{-0.44cm}
     \begin{subfigure}[b]{0.26\textwidth}
         \centering
         \includegraphics[width=\textwidth]{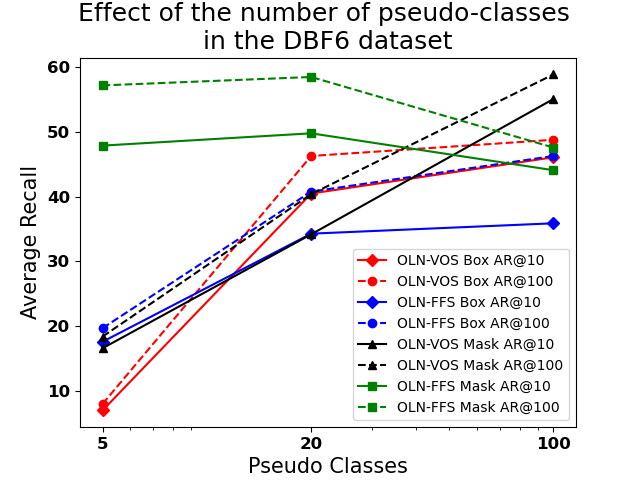}
         \caption[]{}
         \label{fig:dbf6_ablations}
     \end{subfigure}
     \hspace{-0.44cm}
     \begin{subfigure}[b]{0.26\textwidth}
         \centering
         \includegraphics[width=\textwidth]{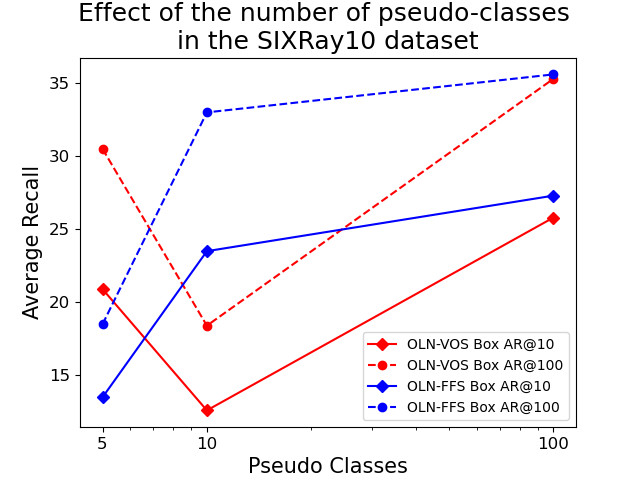}
         \caption[]{}
         \label{fig:sixray10_ablations}
     \end{subfigure}
     \hspace{-0.44cm}
     \begin{subfigure}[b]{0.26\textwidth}
         \centering
         \includegraphics[width=\textwidth]{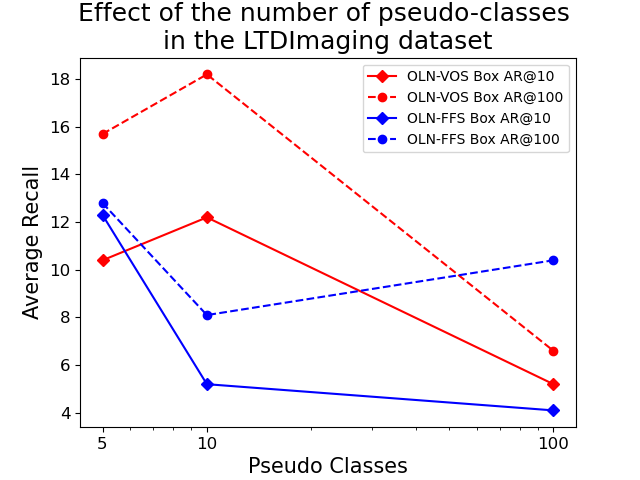}
         \caption[]{}
         \label{fig:ltdimaging_ablations}
     \end{subfigure}
        \caption{Ablation Studies}
        \label{fig:ablations}
\end{figure}

\subsection{Ablation Studies}
\label{sec:ablations}
\cref{fig:ablations} shows the effect of varying the hyperparameters in OLN-SSOS. We present AR@10 and AR@100 metrics when varying the sampling size for SSOS and the number of pseudo-labels. \cref{fig:voc_coco_ablations} shows the effect of the outlier sampling size for the VOC/COCO dataset, ranging from $50$ to $10,000$. A peak in performance is observed for $300$ samples, identifying the boundary region between normal and abnormal samples. While relatively good performance is obtained for all other sampling sizes, it is crucial to identify the optimum choice for each dataset. \cref{fig:dbf6_ablations,fig:sixray10_ablations,fig:ltdimaging_ablations} show the effect of changing the number of pseudo-labels for the DBF6, SIXRay10 and LTDImaging datasets. In general, the higher numbers of pseudo-labels have a positive impact on performance (similar to DeepCluster \cite{caron2018deep}). This suggests the UPC may capture different modalities of data, indicating that over-segmentation helps in anomaly detection. In addition, this offers greater flexibility when designing OoD detectors since we are not tied by the number of ground truth classes. On the other hand, \cref{fig:ltdimaging_ablations} shows a decrease in performance for a higher number of pseudo-classes on the LTDImaging dataset. Differently from the rest of the datasets, LTDImaging only contains three in-distribution classes (person, motorcycle and bicycle) with low inter-class variability, indicating that using several pseudo-classes might cause overfitting. Overall, the number of pseudo-classes must follow a detailed analysis of the dataset to achieve maximal performance. More ablations are available in the Supp. Material.

\section{Conclusion} \label{s:conclude}
\noindent
In this work, we introduce OLN-SSOS, an end-to-end open-world object-based anomaly detection, operating without class supervision to localise unseen anomalies within the training sets. Our method utilises an open-world object detector that learns object pseudo-labels, fitting object features into pseudo-class-conditional Gaussians to synthesise outliers from low-likelihood regions, enabling a better decision boundary between inliers and outliers during inference. We demonstrate the superiority of our approach over baseline methods, which solely rely on existing class-wise data for training in-distribution data. 

Furthermore, we evaluate OLN-SSOS detection performance across different imaging modalities to assess its versatility. Our results reveal that while baseline approaches can detect in-distribution data, they struggle with anomaly detection, particularly in test sets where the anomalies are significantly different from the training classes. Conversely, our approach demonstrates its capability by extending to different imaging modalities (X-ray, infrared), showing improved performance in anomaly detection. 
This provides valuable insights into the generalisation capability of our proposed approach across varying imagery characteristics. Additionally, we extend our method to instance segmentation. The quantitative results illustrate the significant impact of using mask information, yielding better performance in detecting unseen anomalies while still maintaining moderately high in-distribution detection, showcasing the extension of our approach in instance segmentation.

\clearpage

{
    \small
    
\begin{small}
\noindent
\textbf{Acknowledgement:} Partially funded under Innovative Research
Call in Explosives and Weapons Detection, 2023 (sponsored under
the UK Government CONTEST strategy in partnership with US Dept.
Homeland Security, Sci. and Tech. Directorate).
\end{small}
\vspace{-0.5cm}
    \bibliographystyle{splncs04}
    \bibliography{main,sup}
}

\newpage
\title{Towards Open-World Object-based Anomaly Detection via Self-Supervised Outlier Synthesis Supplementary Material}
\titlerunning{Anomaly Detection via Self-Supervised Outlier Synthesis - Supp}
\author{Brian K. S. Isaac-Medina$^\star$, Yona Falinie A. Gaus$^\star$, \\ Neelanjan Bhowmik$^\star$, Toby P. Breckon$^{\star, \dagger}$}

\authorrunning{B.K.S. Isaac-Medina et al.}

\institute{Department of \{$^\star$Computer Science, $^\dagger$Engineering\}, Durham University, UK}

\maketitle
\renewcommand\thesection{\Alph{section}}
\setcounter{figure}{5}  
\setcounter{table}{3}
\section{Dataset Details}

This work uses five different datasets for anomaly detection. While these datasets are focused on object detection and instance segmentation, we can adapt them for our anomaly detection task. The details for the dataset splits and their statistics are described next. 

\subsection{PASCAL-VOC 2007/12 and BDD100k}
We use the \textbf{PASCAL-VOC 2007/12} \cite{everingham2010pascal} dataset as in-distribution (normal) data, containing 20 normal object categories (person, bird, cat, cow, dog, horse, sheep, airplane, bicycle, boat, bus, car, motorcycle, train, bottle, chair, dining table, potted plant, couch and tv). The training partition has 16,551 images and 47,223 objects, while the test set has 4,952 images and 14,997 objects. Additionally, we also use the \textbf{BDD100K} \cite{yu2020bdd100k} dataset as the in-distribution data. This dataset consists of 69,863 images and 1,273,707 objects in the training set and 10,000 images and 185,945 object annotations in the test set, spanning across 10 normal classes (pedestrian, rider, car, truck, bus, train, motorcycle, bicycle, traffic light and traffic sign). For both datasets, MS-COCO is used as the OoD dataset removing the images with overlapping in-distribution objects. In this regard, the COCO test partition without VOC objects consists of 930 images and 2,824 annotations, while the COCO test set without BDD100k in-distribution data has 1,880 images and 8,980 object instances. We use the annotation files provided by Du et al. \cite{du2022vos}.

\subsection{Durham Baggage Full Image (DBF6)}
The DBF6 \cite{akcay2018ganomaly} is an X-ray security imagery dataset containing 6 object classes (firearm, firearm part, laptop, camera, knife and ceramic knife). The experiments involving this dataset use an in-distribution training partition without firearms and firearm parts containing 4,588 images and 5,396 objects, an in-distribution testing set with 1,114 images and 1,340 objects, and an OoD test partition of 692 images and 692 objects (with one single firearm/firearm part per image). The DBF6 dataset is the only dataset with manually annotated instance segmentation masks. 

\subsection{SIXRay10}
The \textbf{SIXRay} \cite{miao2019sixray} dataset consists of 1,059,231 X-ray images with 6 prohibited items (gun, knife, wrench, pliers, scissors and hammer), although it only has 5 annotated classes (hammer is not annotated). In this dataset, we consider the gun as an anomaly and trained on the other classes. We use the SIXRay10 subset, containing 10,000 images, resulting in a training partition with 7,496 normal images and 11,116 objects, and a testing in-distribution partition of 988 images and 1,422 anomaly instances. The OoD test set consists of 352 images with 553 anomalies (guns). 

\subsection{LTDImaging}
Finally, the \textbf{LTDImaging} \cite{nikolov2021seasons} is an infrared video-surveillance dataset with four classes (person, bicycle, motorcycle and vehicle). We train on a week's worth of data considering the vehicle class as an anomaly, giving a training set of 10,108 images and 50,924 objects. For testing, we use a one-day partition (outside the training week) without vehicles, giving a total of 1,570 images and 12,214 objects. For testing OoD detection, we use data from the same training week containing only vehicles (since their occurrence is small compared with the other classes), having a test set of 284 images and 298 objects.

\section{Pseudo-code}

The pseudo-code for training OLN-SSOS/FFS is given in \cref{alg:pseudocode}, while the inference pipeline is described in \cref{alg:inference-pseudocode}.
\begin{algorithm}
\KwData{Input images $\{\mathbf{x}\}_{i=1}^N$, ground truth $\{y_{ij}\}_{i=1,j=1}^{N,N_i}$, where $N_i$ is the number of objects in $\mathbf{x}_i$, pseudo-labels $K$ and OLN-SSOS model $M$.}
\BlankLine
\Begin{

Randomly initialise the pseudo-label centres
$\mathbf{p} \leftarrow \mathcal{N}(\mathbf{0}, I)$
\BlankLine
\ForEach{$\mathit{epoch}$}{
\textcolor{red}{\emph{Ground truth boxes clustering}}\\
$\mathit{fts}\leftarrow\mathit{List}$ \tcc{Object features list}
\For{$i \leftarrow 1$ \KwTo $N$}{
    $\mathbf{f}_i = M.\mathit{Backbone}(\mathbf{x}_i)$\\
    \For{$j \leftarrow 1$ \KwTo $N_i$}{
    $\mathbf{z}_{ij} = \mathit{RoIAlign}(\mathbf{f}_i, y_{ij})$\\
    $\mathit{fts}.\mathrm{append}(\mathbf{z}_{ij})$
    }
}
Initialise kmeans with $\mathbf{p}$.\\
$\mathit{kmeans}=\mathrm{MiniBatchKMeans}(\mathrm{initial}=\mathbf{p})$\\
$\mathit{kmeans}.fit(\mathit{fts})$\\
$\mathbf{p}=\mathit{kmeans}.centres$\\
$\mathbf{c}=\mathit{kmeans}.labels$\tcc{Assigned pseudolabels}
\BlankLine

\textcolor{red}{\emph{Model Training}}\\
\For{$i \leftarrow 1$ \KwTo $N$}{
Extract image features\\
$\mathbf{f}_i \leftarrow M.\mathit{Backbone}(\mathbf{x}_i)$\\
Get Proposals \\
$P_i \leftarrow M.\mathit{RPN}(\mathbf{f}_i)$\\
Pool proposal features \\
$\mathbf{u}_{i} \leftarrow \mathit{RoIAlign}(\mathbf{f}_i, P_i)$ \\
$\mathbf{v}_{i} \leftarrow M.g(\mathbf{u}_i)$ \tcc{Object features. g: Shared head in \cref{fig:full_arch_}}
Predict Pseudo-classes \\
$l_i \leftarrow \mathrm{MLP}(\mathbf{v}_i) $

\BlankLine

\textbf{Outlier Synthesis} \\
$\tilde{\mathcal{V}} \leftarrow \mathit{List}$ \\
$\boldsymbol{\Sigma} \leftarrow \mathit{Cov}(\{\mathbf{v}_i\}, \{l_i\})$\tcc{From \cref{eq:std}}
\For{$k \leftarrow 1$ \KwTo $K$}{
    $\mathcal{V}^{(k)} \leftarrow \{\mathbf{v}_i | \forall (\mathbf{v}_i, l_i), l_i = k \}$\\
    $\boldsymbol{\mu}_k \leftarrow \mathit{Mean}(\mathcal{V}^{(k)})$ \tcc{From \cref{eq:mean}}
    $G_k \leftarrow \mathcal{N}(\boldsymbol{\mu}_k, \boldsymbol{\Sigma})$ \\ 
    Sample virtual outliers \\
    $\mathbf{\tilde{v}}_k \leftarrow \{\mathbf{\tilde{v}}_j | p(\mathbf{\tilde{v}}_j \sim G_k) < \epsilon\}$\\ $\tilde{\mathcal{V}}.append(\mathbf{\tilde{v}}_k)$
}
Get Energies \\
$E_i \leftarrow \mathit{Energy}(\{\mathbf{v}_i\})$ \tcc{Normal energies, from \cref{eq:energy}}
$\tilde{E} \leftarrow \mathit{Energy}(\tilde{\mathcal{V}})$ \tcc{Abnormal energies, from \cref{eq:energy}}
Predict uncertainty score using the anomaly MLP $\phi$\\
$\lambda_i \leftarrow \phi(E_i)$  \\
$\tilde{\lambda} \leftarrow \phi(\tilde{E})$ 
\BlankLine
Get loss values from \cref{eq:rpn-loss,eq:bbox-loss,eq:anomaly-loss}. For FFS, use \cref{eq:nll-loss}. For mask versions, use the error functions \\from Mask R-CNN \cite{he2017mask} and Mask Scoring \cite{huang2019mask} \\
$\mathcal{L} = \mathcal{L}_\mathit{RPN} + \mathcal{L}_\mathit{bbox} + \mathcal{L}_\mathit{mask} + \alpha \mathcal{L}_\mathit{pcls} + \beta \mathcal{L}_\mathit{anomaly} + \gamma \mathcal{L}_\mathit{nll}$
}}}
\caption{OLN-SSOS/FFS Train Pipeline.}\label{alg:pseudocode}
\end{algorithm}

\begin{algorithm}
\KwData{Input image $\{\mathbf{x}\}$ and OLN-SSOS model $M$.}
\BlankLine
\Begin{
\BlankLine
Extract image features\\
$\mathbf{f} \leftarrow M.\mathit{Backbone}(\mathbf{x})$\\
Get Proposals \\
$P \leftarrow M.\mathit{RPN}(\mathbf{f})$\\
Pool proposal features \\
$\mathbf{u} \leftarrow \mathit{RoIAlign}(\mathbf{f}, P)$ \\
$\mathbf{v} \leftarrow M.g(\mathbf{u})$ \tcc{Object features. g: Shared head in \cref{fig:full_arch_}}
\BlankLine
Get object energies \\
$E \leftarrow \mathit{Energy}(\mathbf{v})$ \tcc{From \cref{eq:energy}} 
Predict uncertainty score using the anomaly MLP $\phi$\\
$\lambda \leftarrow \phi(E)$ \\
All objects below a threshold uncertainty score are anomalies.
}
\caption{OLN-SSOS/FFS Inference Pipeline.}
\label{alg:inference-pseudocode}
\end{algorithm}

\section{Training Regime}
We train our models using the MMDetection \cite{mmdetection} framework, with slight variations for each dataset. For all of our datasets, except BDD100K, we initialize our models with the OLN \cite{kim2021oln} pre-trained on the VOC dataset, as per the original implementation (see \cite{kim2021oln} for the details). Since some classes in VOC are considered anomalies for the BDD100K experiment, we trained an OLN on the BDD100k dataset and used it to initialize the BDD100K experiments. For all the experiments except the LTDImaging, we resize the images to have a maximum side length of 1,333 pixels and variable minimum side length to allow for multi-scale training. Since LTDImaging images come from the same camera, we kept the same image size for training and testing, \ie,  384 $\times$ 288 pixels. For all the experiments, random horizontal flip is used during training and 0-padding is added so the images are exactly divisible by 32. 

For all our experiments, OLN-SSOS and OLN-FFS are trained for 8 epochs with an initial learning rate of 0.001 with a linear warmup for the first 100 iterations, decaying by a factor of 10 after epoch 4. All our training is carried out using stochastic gradient descent with a weight decay of $1\times10^{-4}$. Considering the pertaining time of OLN, our models are trained for a similar number of epochs as in VOS \cite{du2022vos} (in total, we train for 16 epochs while VOS is trained for $\sim18$ epochs). All our models are trained with a batch size of 2 except for LTDImaging, which uses a batch size of 64. Pseudo-class training starts from the beginning, reclustering before each epoch.

\section{Analysis of learned pseudo-classes}
\begin{figure}
\centering
\subfloat{\includegraphics[width=\linewidth]{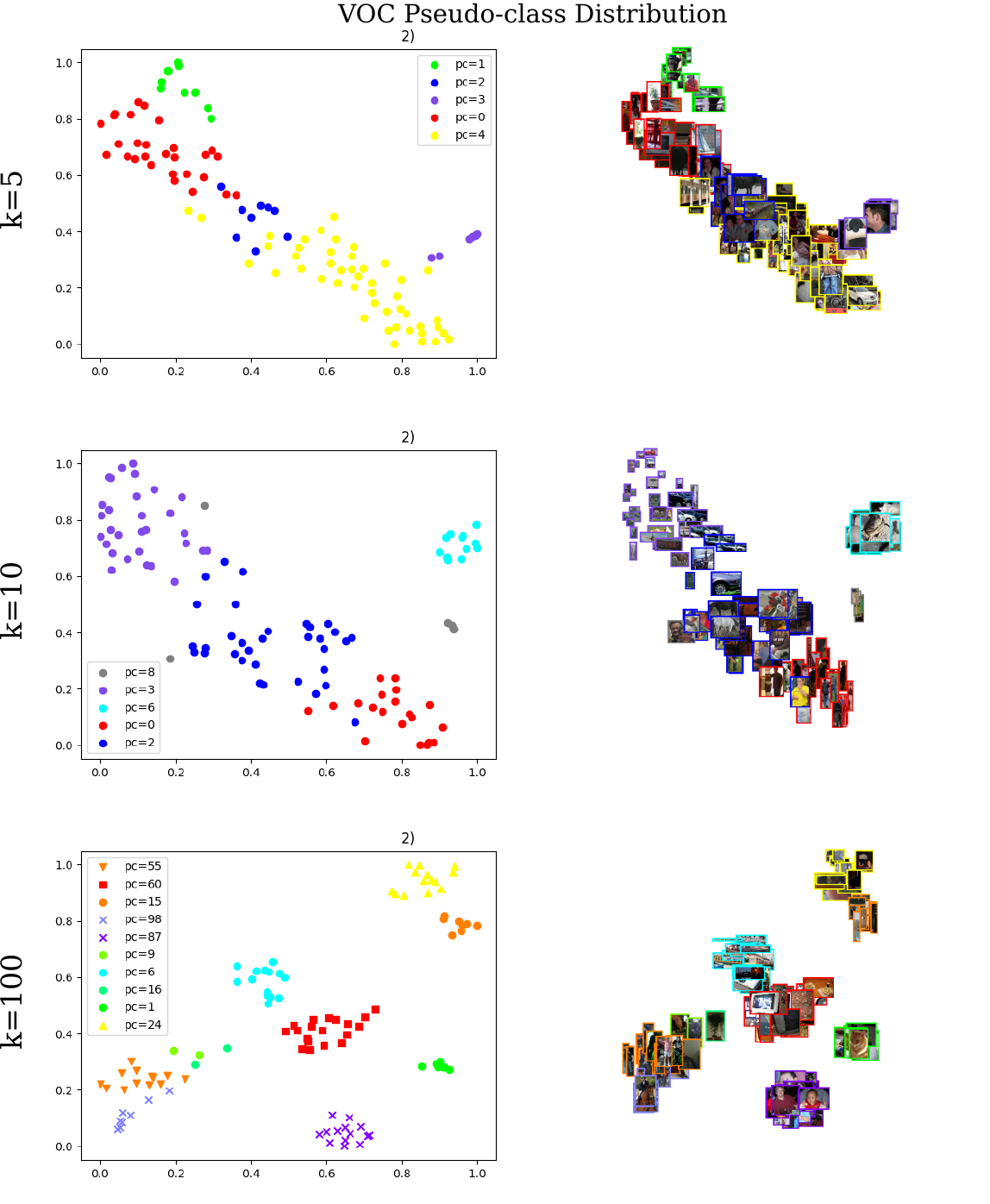}}
\caption{Learned pseudo-labels for the VOC dataset.}  
\label{fig:voc_pseudo_labels}
\end{figure}

\begin{figure}
\centering
\subfloat{\includegraphics[width=\linewidth]{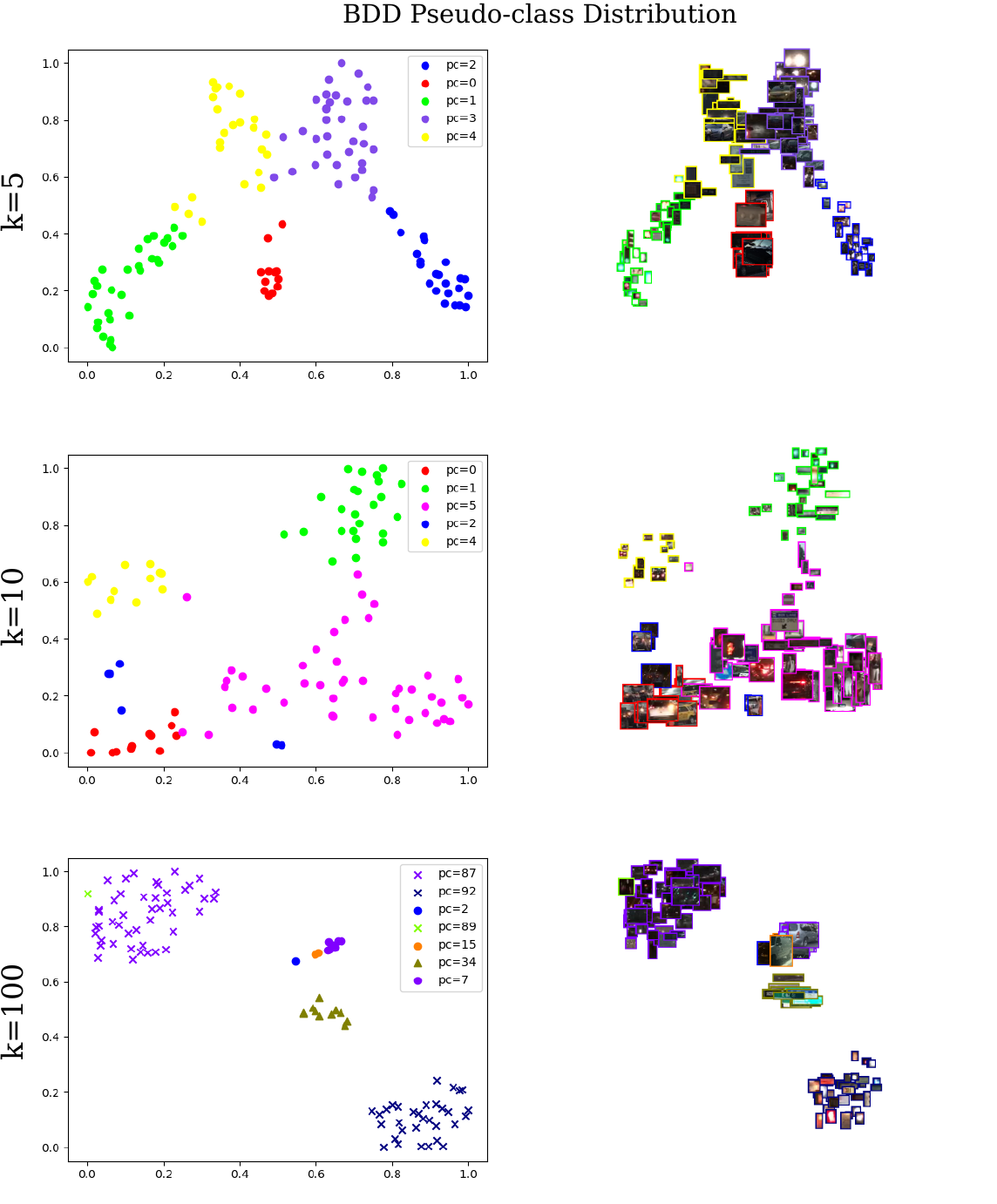}}
\caption{Learned pseudo-labels for the BDD dataset.}  
\label{fig:bdd_pseudo_labels}
\end{figure}

\begin{figure}
\centering
\subfloat{\includegraphics[width=\linewidth]{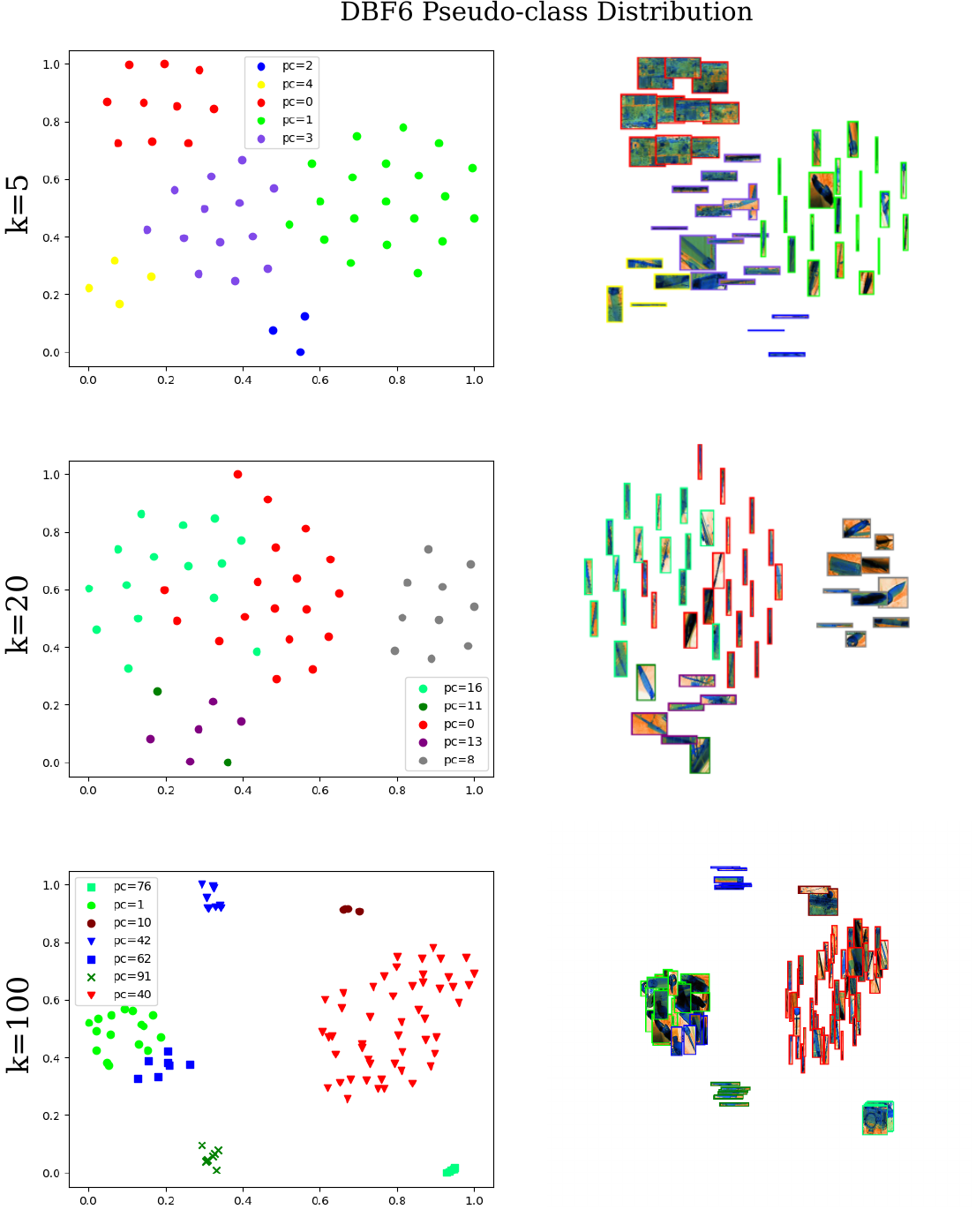}}
\caption{Learned pseudo-labels for the DBF6 dataset.}  
\label{fig:dbf6_pseudo_labels}
\end{figure}

\begin{figure}
\centering
\subfloat{\includegraphics[width=0.85\linewidth]{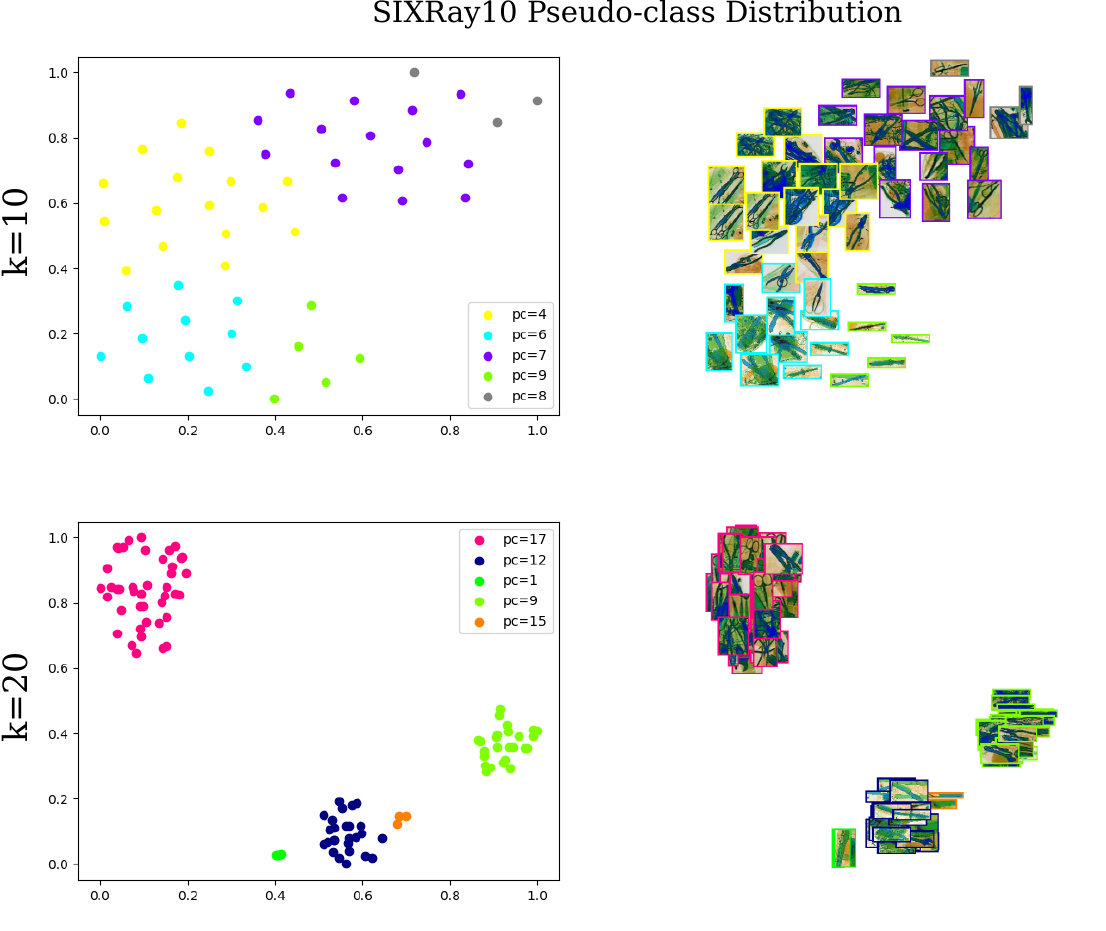}}
\caption{Learned pseudo-labels for the SIXRay10 dataset.}  
\label{fig:sixray10_pseudo_labels}
\end{figure}

\begin{figure}[t]
\centering
\subfloat{\includegraphics[width=0.85\linewidth]{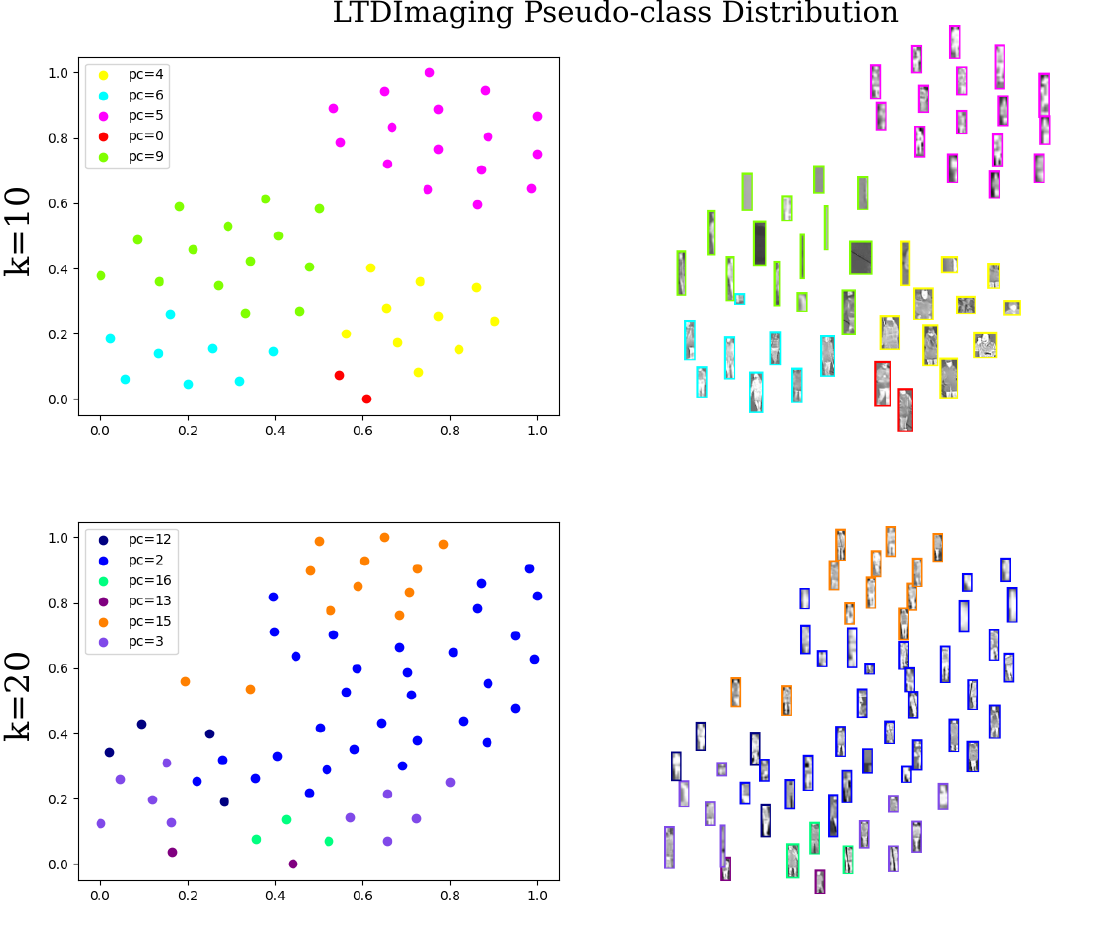}}
\caption{Learned pseudo-labels for the LTDImaging dataset.}  
\label{fig:ltd_pseudo_labels}
\end{figure}

\cref{fig:voc_pseudo_labels,fig:bdd_pseudo_labels,fig:dbf6_pseudo_labels,fig:sixray10_pseudo_labels,fig:ltd_pseudo_labels} show a t-SNE \cite{van2008visualizing} projection of the object features, clustered by their learned pseudo-classes. It is seen in \cref{fig:voc_pseudo_labels} that the VOC dataset is not properly clustered when using only a few pseudo-labels. For instance, for pseudo-labels $k=5$ and $k=10$, no significant difference can be observed among the pseudo-labels, specially when clustering people instances. Although some semantic separation can be observed between vehicles and people, there is still some confusion for $k=10$. On the other hand, when using a large number of pseudo-classes, such as $k=100$, it is seen that the learned labels dive the objects into more semantically meaningful clusters, such as animals (green), seated people (purple) or indoor objects (red). However, this clustering is still challenging and it demonstrates why our method does not match the state of the art for this dataset. A similar trend is observed in \cref{fig:bdd_pseudo_labels} for the BDD dataset, although the pseudo-labels cluster the objects better for smaller $k$ compared with VOC. Pseudo-clusters for the DBF6 dataset is shown in \cref{fig:dbf6_pseudo_labels}. Given the more balanced distribution of categories, clusters seem to capture semantically similar objects, even for a small number of pseudo classes. It is also observed that our method seems to differentiate between different orientations of knives, while keeping all laptops in a similar class. Additional results for X-Ray imagery is presented for the SIXRay10 dataset in \cref{fig:sixray10_pseudo_labels}, showing that while some objects might look similar (scissors and tweezers), they can still be separated into different classes with no class training. Finally, \cref{fig:ltd_pseudo_labels} shows the pseudo-label analysis for the LTDImaging dataset. While some semantic separation can be observed (for instance, people in similar poses are clustered together), the low object variability makes it difficult to separate them into meaningful clusters, meaning that over-segmentation might negatively impact the performance, as seen in \cref{fig:ablations}.

\section{Comparison against other open-world detectors}

While OLN-SSOS focuses on localising objects and labelling them as anomalies, our goal can be seen as  similar to OW-DETR \cite{gupta2022ow} and PROB \cite{Zohar_2023_CVPR}. Therefore, we include a comparison of our approach against such open-world object detectors in
\cref{tab:prob-comp}. We compare PROB with the best results for each dataset in our work (excluding BDD). Unknown recall at 0.5 IoU (UR\textsubscript{0.5}) and COCO AR@100 are reported. Both metrics are for 100 detections (PROB uses $100$ object queries). A similar trend is observed, i.e., it performs better in VOC/COCO (our method being competitive) but our method is superior in the other tasks. It is worth noting that PROB also uses class supervision to detect unknown classes. Specifically, an unknown object is detected if they have a high objectness score but a low known class probability. Also, our evaluation results trend mirror the single task for OW-DETR/PROB. These results further showcase the ability of our method for anomaly detection and localisation without class supervision.

\begin{table}[t]
\caption{\scriptsize PROB \cite{Zohar_2023_CVPR} vs Ours.}
\label{tab:prob-comp}
\resizebox{\linewidth}{!}{%
\begin{tabular}{|l|cc|cc|cc|cc|}
\hline
\multirow{2}{*}{} & \multicolumn{2}{c|}{VOC/COCO} & \multicolumn{2}{c|}{DBF6}  & \multicolumn{2}{c|}{SIXRay10} & \multicolumn{2}{c|}{LTD} \\ \cline{2-9} 
                  & \multicolumn{1}{c|}{UR\textsubscript{0.5}}  & AR@100 & \multicolumn{1}{c|}{UR\textsubscript{0.5}} & AR@100 & \multicolumn{1}{c|}{UR\textsubscript{0.5}} & AR@100 & \multicolumn{1}{c|}{UR\textsubscript{0.5}} & AR@100 \\ \hline
PROB              & \multicolumn{1}{c|}{\textbf{53.2}}  & \textbf{34.0}  & \multicolumn{1}{c|}{26.9} & 6.2  & \multicolumn{1}{c|}{57.5} & 13.0  & \multicolumn{1}{c|}{3.29} & 1.0  \\ \hline
Ours              & \multicolumn{1}{c|}{40.4}        & 17.8 & \multicolumn{1}{c|}{\textbf{90.5}}        & \textbf{48.8}  & \multicolumn{1}{c|}{\textbf{92.4}}        & \textbf{35.6}  & \multicolumn{1}{c|}{\textbf{38.6}}  & \textbf{18.2}  \\ \hline
\end{tabular}%
}
\end{table}

\section{Further Ablations}
\begin{figure}
\centering
\subfloat{\includegraphics[width=0.6\linewidth]{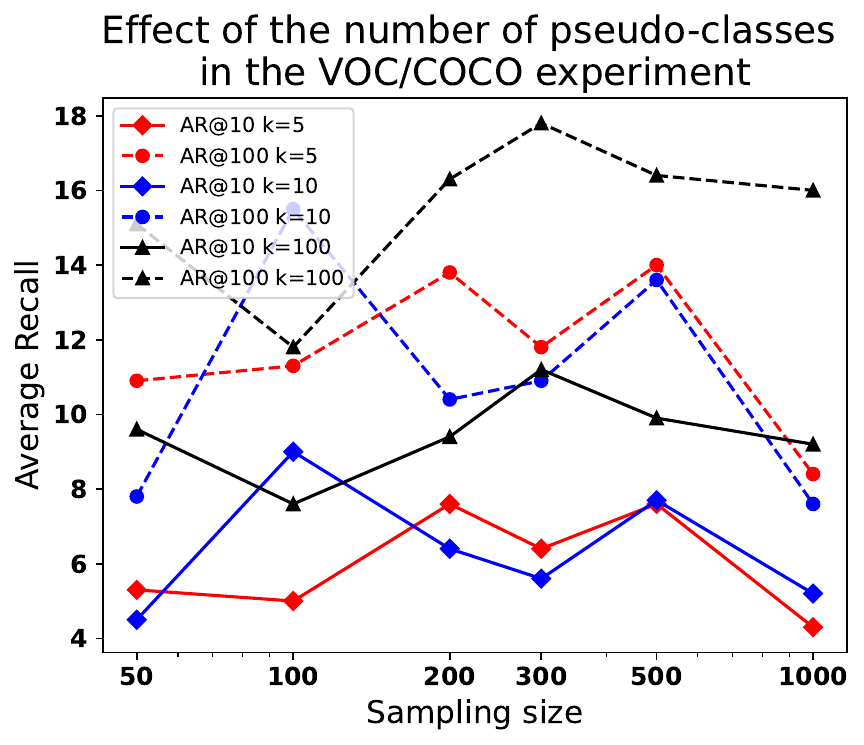}}

\caption{Ablations for VOC/COCO. Maximum performance is achieved for k = 100 and 300 samplings for virtual outlier synthesis.}  
\label{fig:voccoco_ablations}
\end{figure}

\begin{figure}
\centering
\begin{minipage}{.45\textwidth}
  \centering
  \includegraphics[width=\linewidth]{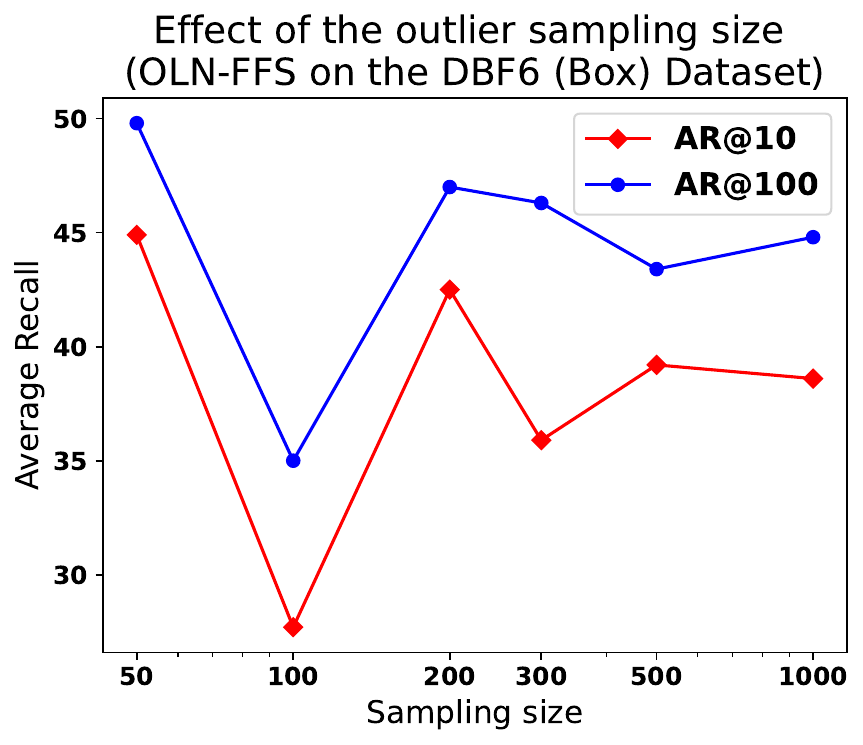}
  \captionof{figure}{Ablations for DBF6 (Box).}
  \label{fig:dbf6box_ablations}
\end{minipage}%
\begin{minipage}{.45\textwidth}
  \centering
  \includegraphics[width=\linewidth]{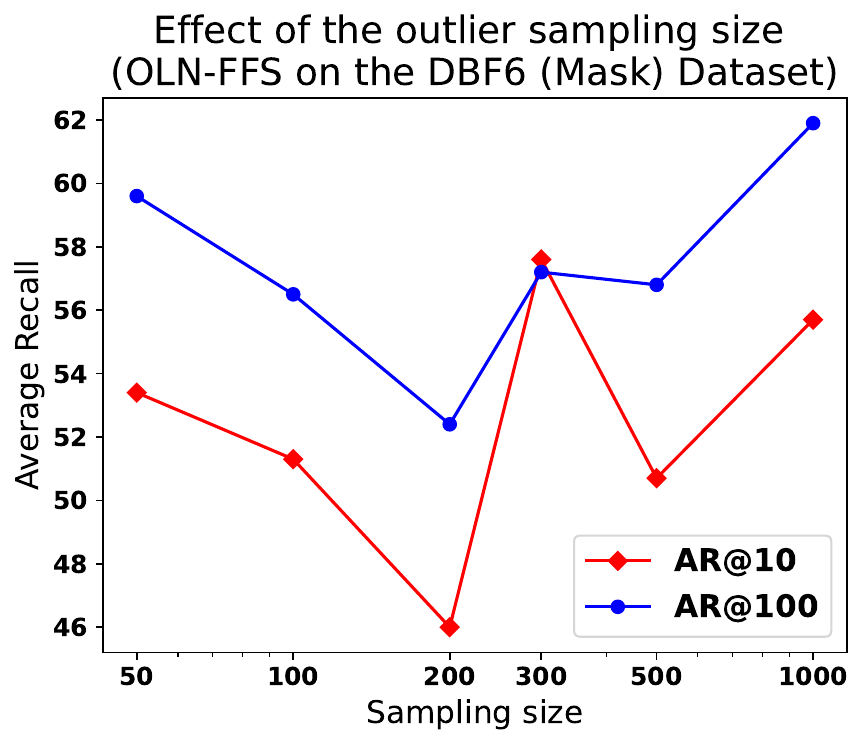}
  \captionof{figure}{Ablations for DBF6 (Mask).}
  \label{fig:db6mask_ablations}
\end{minipage}
\end{figure}

\begin{figure}
\centering
\begin{minipage}{.45\textwidth}
  \centering
  \includegraphics[width=\linewidth]{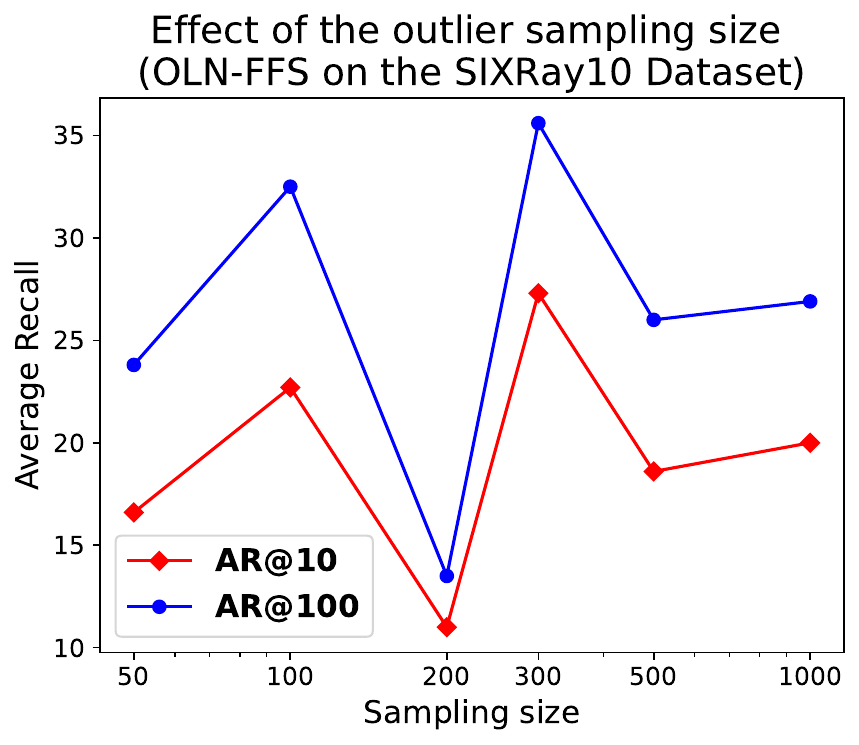}
  \captionof{figure}{Ablations for SIXRay10.}
  \label{fig:sixray10_ablations_supp}
\end{minipage}%
\begin{minipage}{.45\textwidth}
  \centering
  \includegraphics[width=\linewidth]{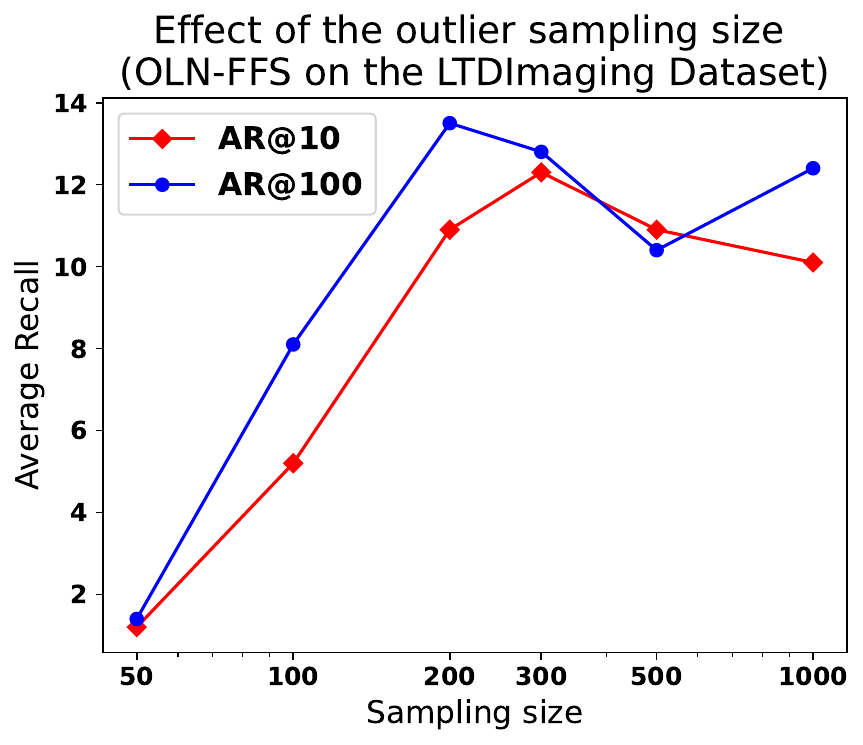}
  \captionof{figure}{Ablations for LTDImaging.}
  \label{fig:ltdimaging_ablations_supp}
\end{minipage}
\end{figure}

\cref{fig:voccoco_ablations,fig:dbf6box_ablations,fig:db6mask_ablations,fig:sixray10_ablations_supp,fig:ltdimaging_ablations_supp} show further ablation studies for varying sampling sizes. In particular, \cref{fig:voccoco_ablations} show the average recall for different sampling sizes and different numbers of pseudo-classes; \cref{fig:dbf6box_ablations,fig:db6mask_ablations} show the ablations for the OoD sampling size for DBF6 Box and DBF6 Mask; and \cref{fig:sixray10_ablations_supp,fig:ltdimaging_ablations_supp} show the ablations for SXIRay10 and LTDImaging. While the best results for VOC/COCO are obtained for a sampling size of 300 images and with 100 pseudo-classes, being the reason why we used this sampling size in our experiments, this might not be the same for different datasets, therefore indicating that a proper evaluation must be carried out for each dataset.

\section{Qualitative Results}

\begin{figure*}[t!]
\centering
\subfloat{\includegraphics[width=0.8\linewidth]{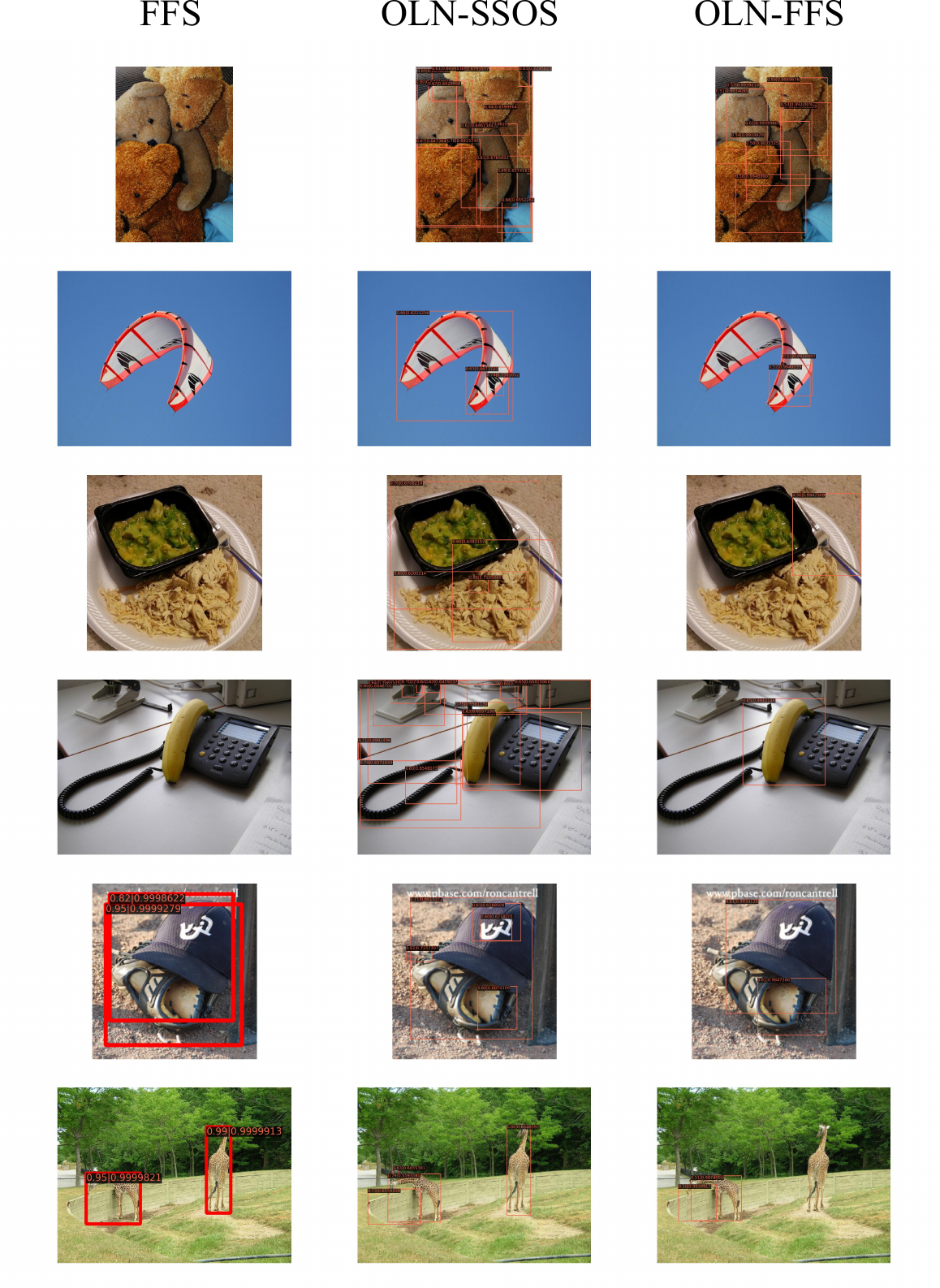}}
\caption{Qualitative results for VOC/COCO. It is observed that while OLN-SSOS gets more objects, OLN-FFS can get objects with more quality. In some instances. OLN-FFS misses objects of interest. It is also observed that FFS only detect objects closer to the training set, like animals or a cap (similar to a human head).}  
\label{fig:voccoco_qual}
\end{figure*}

\begin{figure*}[t!]
\centering
\subfloat{\includegraphics[width=\linewidth]{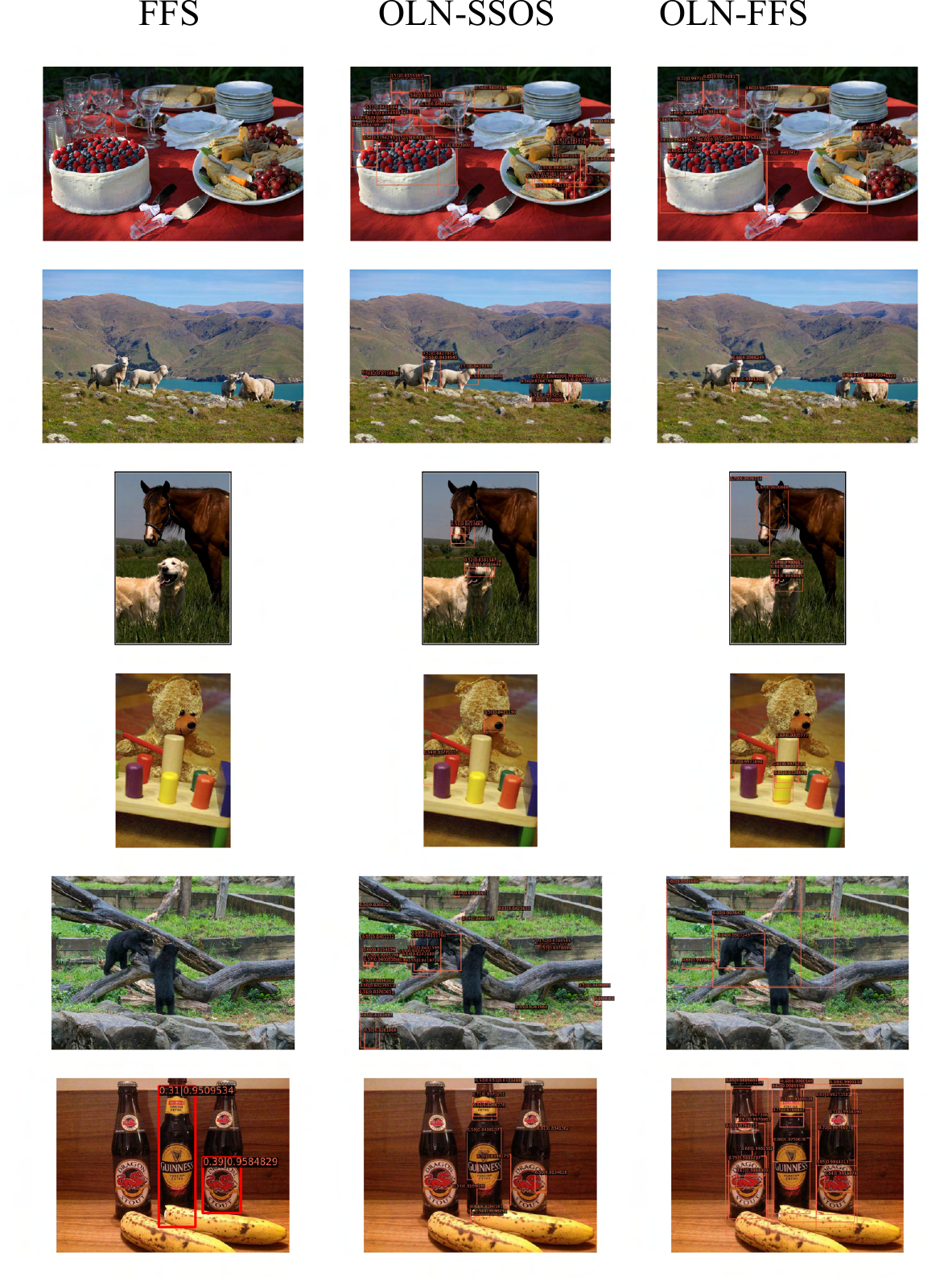}}
\caption{Qualitative results for BDD/COCO. Similar to VOC/COCO, OLN-FFS gets less objects but with more quality (see the last row). In this example, FFS gets less images since it has fewer training classes.}  
\label{fig:bdd_qual}
\end{figure*}

\begin{figure*}[t!]
\centering
\subfloat{\includegraphics[width=\linewidth]{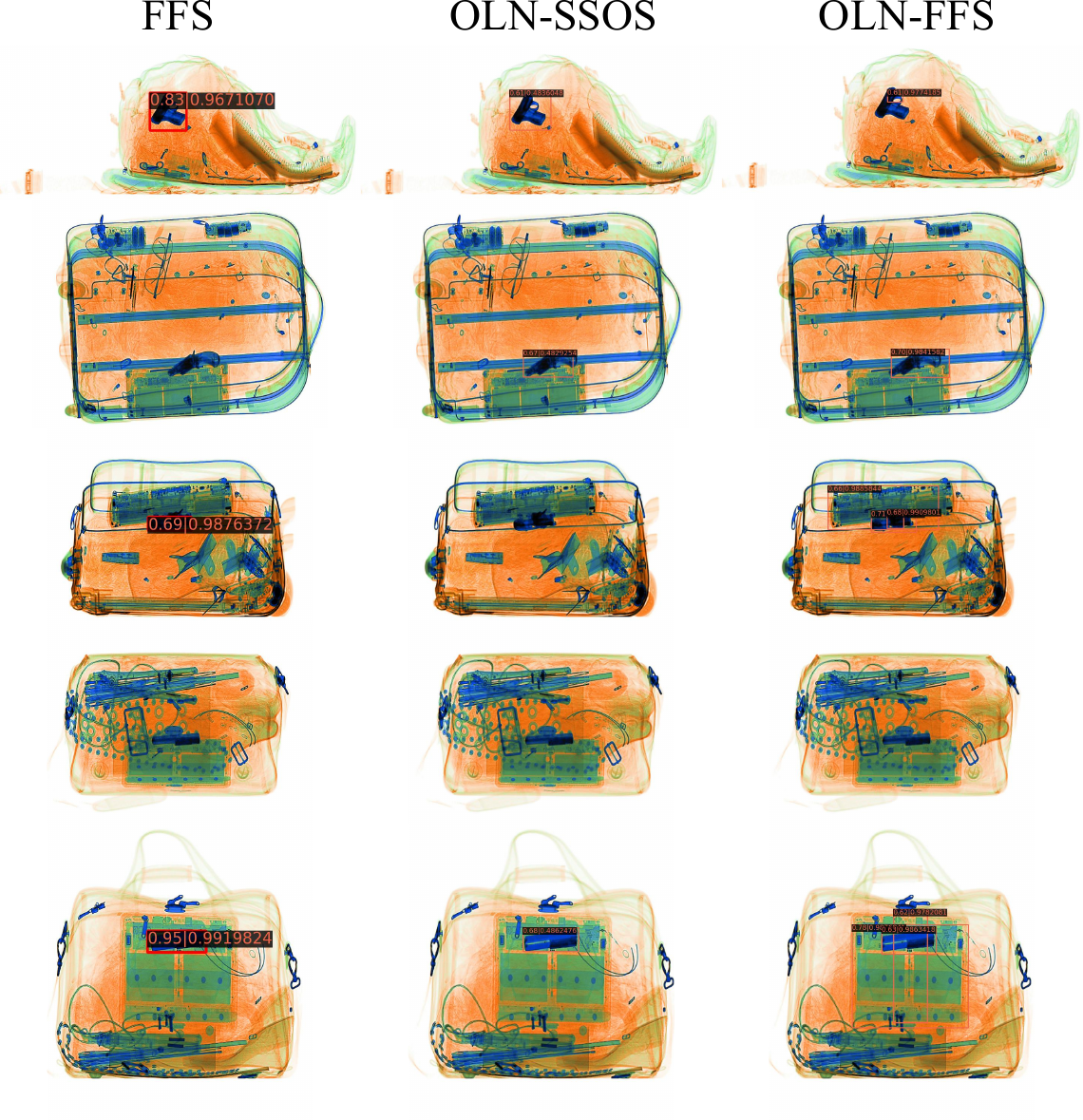}}
\caption{Qualitative results for DBF6 (Box). While FFS has a relative good performance, it sometimes misses objects like the firearm in the second row. Additionally, OLN-FFS detects other anomalies that are not in the test set, like the tablet in the fourth row. There are some cases where none of the models can detect the anomaly, like the fifth row.}  
\label{fig:dbf6box_qual}
\end{figure*}

\begin{figure*}[t!]
\centering
\subfloat{\includegraphics[width=\linewidth]{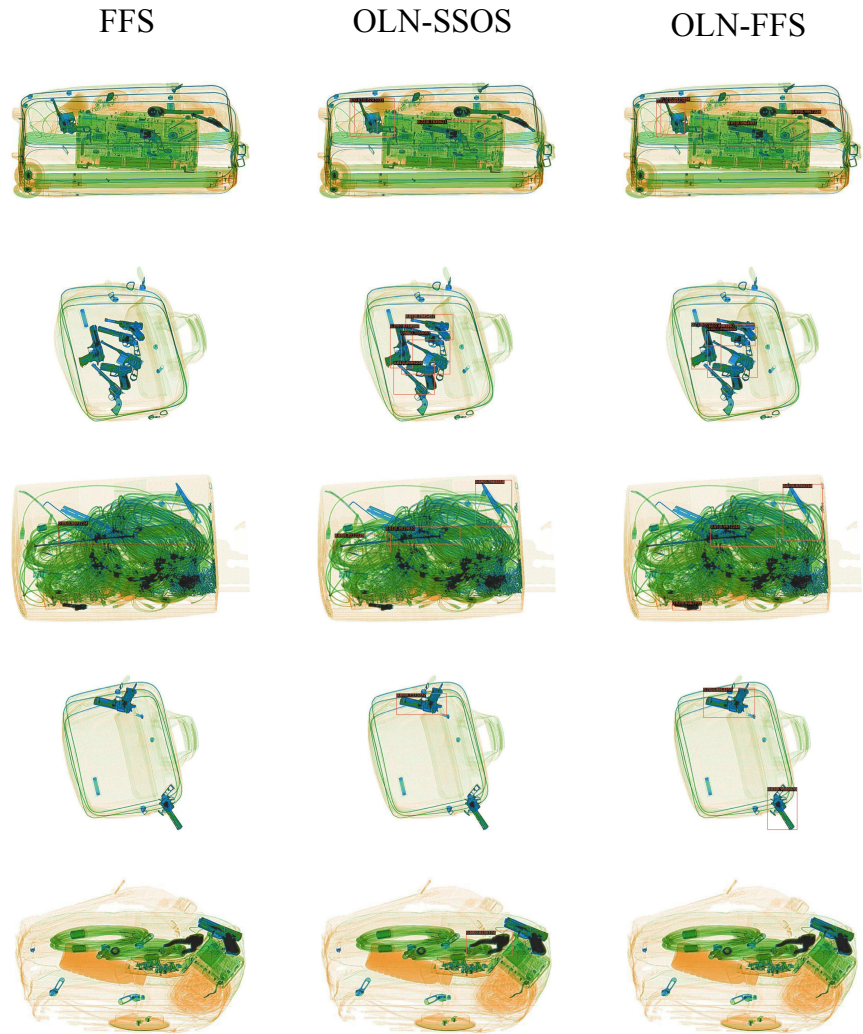}}
\caption{Qualitative results for SIXRay10. In all of the examples, FFS misses the anomaly, while OLN-SSOS and OLN-FFS get most of the anomalies.}  
\label{fig:sixray_qual}
\end{figure*}

\begin{figure*}[t!]
\centering
\subfloat{\includegraphics[width=\linewidth]{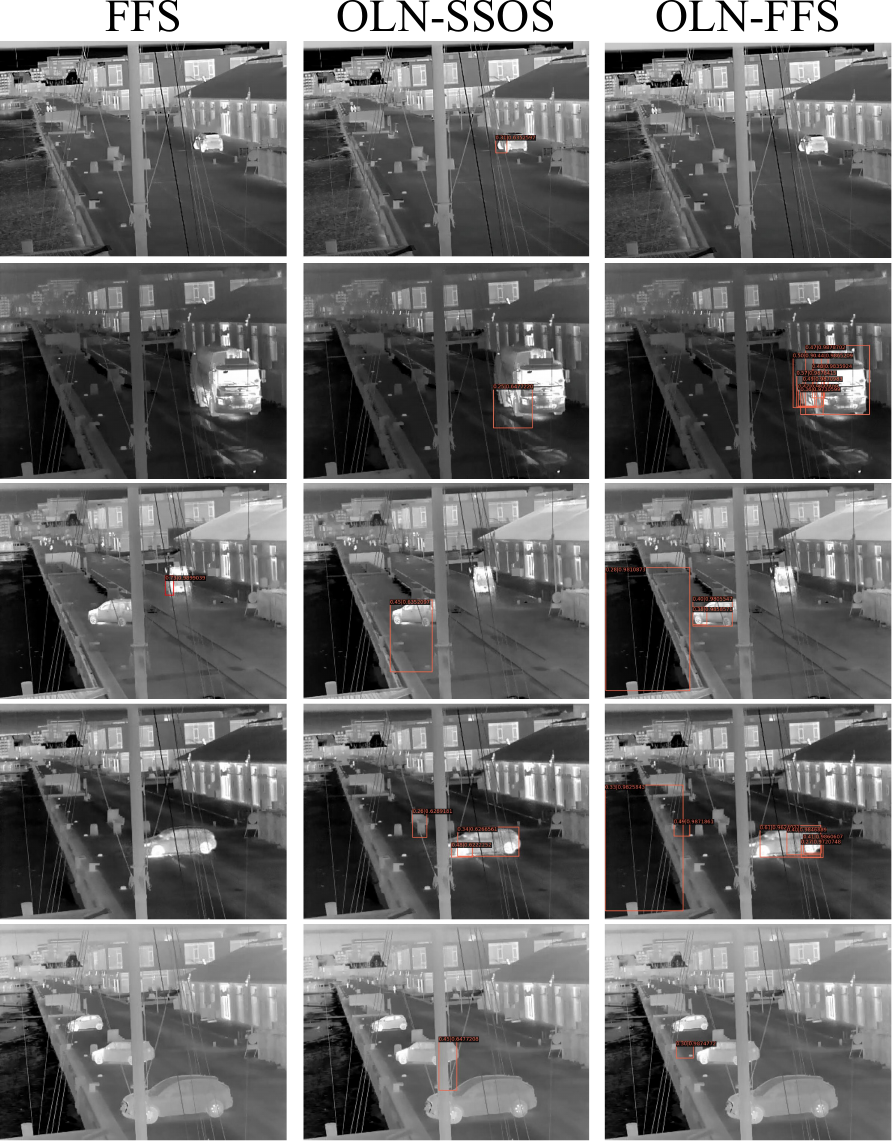}}
\caption{Qualitative results for LTDImaging. Similar to SIXRay10, FFS misses almost all the anomalies. It can be seen that OLN-FFS detects more anomalies than OLN-SSOS, although they both fail in some instances (last row).}  
\label{fig:ltd_qual}
\end{figure*}

\begin{figure*}[t!]
\centering
\subfloat{\includegraphics[width=\linewidth]{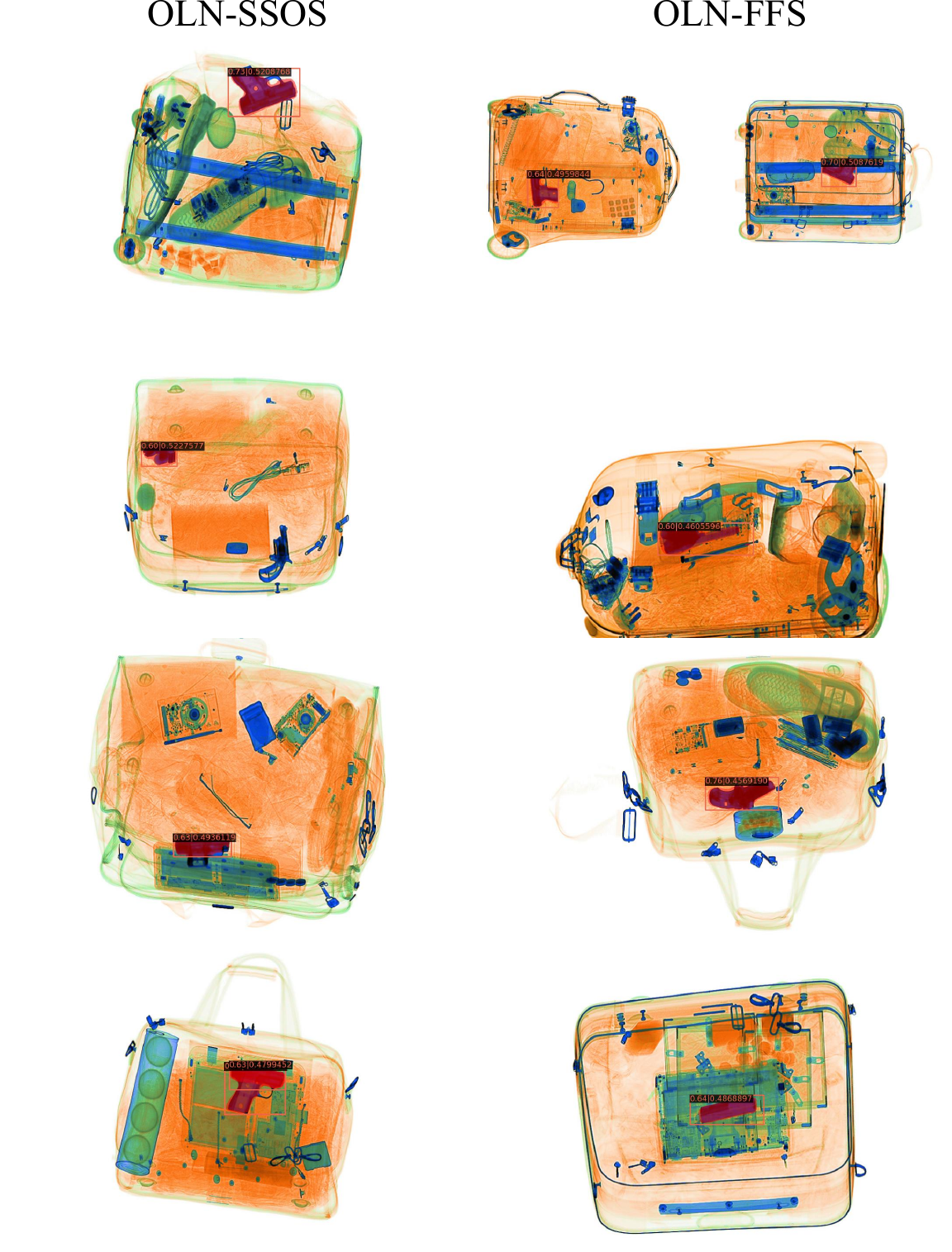}}
\caption{Qualitative results for DBF6 (Mask). No baseline is presented since there is no baseline for instance segmentation. In all of the examples, it can be noted that both methodologies get the correct segmentation mask, with the exception of the missed gun for OLN-SSOS in the second row.}  
\label{fig:dbf6mask_qual}
\end{figure*}

\cref{fig:voccoco_qual,fig:bdd_qual,fig:dbf6box_qual,fig:sixray_qual,fig:ltd_qual} show more qualitative results for the bounding box models (only the baseline FFS \cite{kumar2023normalizing} is included). \cref{fig:dbf6mask_qual} shows additional qualitative results for our mask models. 

\end{document}